\documentclass{article}
\usepackage{array}
\usepackage{booktabs}
\usepackage{graphicx}
\usepackage{adjustbox}
\usepackage{amssymb}
\usepackage{xcolor}
\usepackage{pifont}

\usepackage{array}
\usepackage{float}
\usepackage{makecell}
\usepackage{amsmath}
\usepackage{multirow}  
\definecolor{darkgreen}{RGB}{0,100,0} % 
\definecolor{ModelGreen}{RGB}{215,230,212}
\usepackage{enumitem}
\usepackage{booktabs}       % \toprule, \midrule, \bottomrule
\usepackage{multirow}       % \multirow
\usepackage{makecell}       %
\usepackage{colortbl}       %
\usepackage{graphicx}       
\usepackage{adjustbox}      
\usepackage{caption} 
\usepackage{enumitem}
\usepackage{float}

\usepackage[hidelinks,colorlinks=true,urlcolor=black,linkcolor=purple,citecolor=Highlight]{hyperref}

    \PassOptionsToPackage{numbers, compress}{natbib}
\usepackage[preprint] {neurips_2025}

\usepackage[utf8]{inputenc} % allow utf-8 input
\usepackage[T1]{fontenc}    % use 8-bit T1 fonts
\usepackage{hyperref}       % hyperlinks
\usepackage{url}            % simple URL typesetting
\definecolor{Highlight}{HTML}{39a58a} 
\definecolor{darkGreen}{RGB}{92, 148, 110}
\newcommand{\darkGreen}[1]{\textcolor{darkGreen}{#1}}

\title{VidText: Towards Comprehensive Evaluation for Video Text Understanding}

\author{
\textbf{Zhoufaran Yang}\textsuperscript{2,*} \quad
\textbf{Yan Shu}\textsuperscript{1,*} \quad
\textbf{Jing Wang}\textsuperscript{3} \quad
\textbf{Zhifei Yang}\textsuperscript{4,\dag}\quad 
\textbf{Yan Zhang}\textsuperscript{5,6}  \quad
\textbf{Yu Li}\textsuperscript{2} \quad\\
\textbf{Keyang Lu}\textsuperscript{7} \quad
\textbf{Gangyan Zeng}\textsuperscript{8}  \quad 
\textbf{Shaohui Liu}\textsuperscript{2} \quad
\textbf{Yu Zhou}\textsuperscript{9} \quad
\textbf{Nicu Sebe}\textsuperscript{1} \\
$^{1}$UNITN \quad
$^{2}$HIT \quad
$^{3}$SEU \quad
$^{4}$PKU\quad 
$^{5}$IIE, CAS \quad
$^{6}$UCAS \quad
$^{7}$BUAA \quad 
$^{8}$NJUST \quad  
$^{9}$NKU \\
\small $^*$Equal contribution. \\
\small $^\dag$Corresponding author. \\
\href{https://github.com/shuyansy/VidText}{\textcolor{cyan}{\texttt{https://github.com/shuyansy/VidText}}}
}

\begin{document}

\maketitle

\begin{abstract}
Visual texts embedded in videos carry rich semantic information, which is crucial for both holistic video understanding and fine-grained reasoning about local human actions. However, existing video understanding benchmarks largely overlook textual information, while OCR-specific benchmarks are constrained to static images, limiting their ability to capture the interaction between text and dynamic visual contexts.
To address this gap, we propose VidText, a new benchmark designed for comprehensive and in-depth evaluation of video text understanding. VidText offers the following key features: 1) It covers a wide range of real-world scenarios and supports multilingual content, encompassing diverse settings where video text naturally appears. 2) It introduces a hierarchical evaluation framework with video-level, clip-level, and instance-level tasks, enabling assessment of both global summarization and local retrieval capabilities. 3) The benchmark also introduces a set of paired perception reasoning tasks, ranging from visual text perception to cross-modal reasoning between textual and visual information. Extensive experiments on 18 state-of-the-art  Large Multimodal Models (LMMs) reveal that current models struggle across most tasks, with significant room for improvement. Further analysis highlights the impact of both model-intrinsic factors, such as input resolution and OCR capability, and external factors, including the use of auxiliary information and Chain-of-Thought reasoning strategies. We hope VidText will fill the current gap in video understanding benchmarks and serve as a foundation for future research on multimodal reasoning with video text in dynamic environments.

\end{abstract}

\section{Introduction}
\label{sec:introduction}
Large Multimodal Models (LMMs) \cite{bai2025qwen2,chen2024internvl, liu2023visual,alayrac2022flamingo,li2023blip} are rapidly emerging as general-purpose solutions for a wide range of vision-language tasks, demonstrating impressive perception and cognitive capabilities across various multimodal benchmarks. Building on this success, there is a growing interest in extending LMMs to video understanding \cite{zhang2025videollama,li2023videochat,llamavid,song2024moviechat,fang2023alignment,ataallah2024minigpt4}, including video captioning, question-answering and retrieval~\cite{fang2023mask}. To support this development, a number of video benchmarks \cite{fu2024video,zhou2024mlvu,li2024mvbench,wu2024longvideobench,chandrasegaran2024hourvideo} have recently been proposed to enable more comprehensive evaluations of LMMs in dynamic visual environments.

However, existing video understanding evaluations primarily focus on major events, character actions, and interpersonal relationships, while largely overlooking video text. As a self-descriptive visual component, text in videos  plays a crucial role in visual understanding \cite{zhang2025track,zhao2022towards,xu2021videoclip}. On one hand, it provides \textbf{explicit perceptual cues}, such as street signs, storefronts, or subtitles, that help identify key elements and clarify the scene. On the other hand, text also enables \textbf{contextual reasoning}, revealing underlying motivations or causal relationships. For example, a “SALE” sign in a shop may explain why people are gathering, which is not readily apparent from visual cues alone.

Compared to images, perceiving dynamic video text and modeling its interaction with evolving visual contexts in videos is significantly more challenging. It requires not only fine-grained localization at the instance level, but also temporal tracking and spotting at the clip level, as well as holistic understanding at the video level. Furthermore, video text appears in a wide range of scenarios and across multiple languages, which further increases the complexity of recognition and reasoning. Based on these insights, we propose \textbf{VidText}, a comprehensive benchmark for video text understanding, which introduces the following key advantages:

\begin{itemize}[leftmargin=1.2em]
    \item  \textbf{It encompasses a wide variety of video genres}, including media, entertainment, ego-centric, sports, life record and knowledge,  with 27 fine-grained categories covering diverse scenarios rich in visual text, such as scene text and subtitles. Moreover, it includes multilingual content, covering English, Chinese, Korean, Japanese, and German.
\item \textbf{It supports multi-granularity evaluation}, including video-level, clip-level, and instance-level tasks. Video-level tasks involve holistic OCR understanding and reasoning over global video content. Clip-level tasks are designed to require localized comprehension based on specific temporal segments. Instance-level tasks demand fine-grained temporal and spatial grounding of individual text instances to support precise question answering.
\item \textbf{It spans from visual text perception to cross-modal reasoning with visual context}. Building upon the meticulously annotated video text data, we produce video text-centric Chain-of-Thought (CoT) annotations, explicitly capturing the reasoning process between video descriptions and embedded texts, including spatial relationships with surrounding objects and temporal dependencies related to actions or events. In this way, we extend video text perception tasks into their corresponding reasoning counterparts, forming a comprehensive paired perception–reasoning framework that spans eight tasks covering multiple levels of understanding.

\end{itemize}

Tab. \ref{tab:benchmark-comparison} shows that VidText enables a more comprehensive evaluation of video text understanding compared to both general video understanding benchmarks and video text-specific benchmarks. We conduct extensive evaluations on 18 popular LMMs, revealing several important insights. First, \textit{video text understanding remains a technically challenging task for existing models}. Although Gemini 1.5 Pro \cite{team2023gemini} achieves the highest performance, it only reaches an average score of 46.8\%, and all models perform poorly on multi-granularity tasks, which is far below estimated human-level performance. Second, several concurrent open-source models \cite{zhang2025videollama,chen2024expanding} demonstrate competitive performance, narrowing the gap with proprietary systems. Third, our empirical findings highlight several influential factors in video text understanding, including the model's image OCR capability, input resolution, the integration of auxiliary information, and the role of Chain-of-Thought reasoning strategies. Therefore, we believe VidText serves as a valuable complement to existing general video understanding benchmarks, while also providing new insights for the OCR and multimodal reasoning communities.

\begin{table*}[t]
  \centering
    \caption{
    Comparison of video understanding benchmarks. ``Vid'', ``Cli'' and ``Ins'' denote video-level, clip-level and instance-level tasks. ``T'', ``S'' and 
    ``C" mean temporal, spatial and causal dimensions. ``MC'' and ``OE'' represent multiple-choice and open-ended questions. ``TQ'': the percentage of questions containing visual text instances.}
    \vspace{-5pt}
  \renewcommand{\arraystretch}{1.2}
  \begin{adjustbox}{max width=\textwidth}
    \begin{tabular}{
      l
      c c c               % Video  QA  TQ
      c c c               % VideoText  MultiSrc  MultiGran
      c c                 % Percep T S
      c c c               % Reason T S C
      c c                 % Paired  TaskType
    }
    \toprule
    \multirow{2}{*}{\textbf{Benchmark}} &
    \multirow{2}{*}{\textbf{Video}} &
    \multirow{2}{*}{\textbf{QA}} &
    \multirow{2}{*}{\textbf{TQ\,\%}} &
    \multirow{2}{*}{\makecell{\textbf{Multi}\\\textbf{ Lingual}}} &
    \multirow{2}{*}{\makecell{\textbf{Multi}\\\textbf{Source}}} &
    \multirow{2}{*}{\makecell{\textbf{Multi}\\\textbf{Granularity}}} &
    \multicolumn{2}{c}{\textbf{Perception}} &
    \multicolumn{3}{c}{\textbf{Reasoning}} &
    \multirow{2}{*}{\makecell{\textbf{Paired}\\\textbf{Tasks}}} &
    \multirow{2}{*}{\textbf{TaskType}} \\
    \cmidrule(lr){8-9}\cmidrule(lr){10-12}
    & & & & & & & \textbf{T} & \textbf{S} & \textbf{T} & \textbf{S} & \textbf{C} & & \\
    \midrule
    % -------- Block 1 ----------
%    \rowcolor[HTML]{EFEFEF}
%   \multicolumn{14}{l}{\textit{Traditional Video-Spotting Datasets}} \\[-0.2em]

    % -------- Block 2 ----------
    \rowcolor[HTML]{EFEFEF}
    \multicolumn{14}{l}{\textit{General Video Understanding Datasets}} \\
    NExT-QA \cite{xiao2021next} & 5,440 & 52,044 & – & \textcolor{red}{\ding{55}} & \textcolor{darkgreen}{\checkmark} & Vid+Cli
            & \textcolor{red}{\ding{55}} & \textcolor{red}{\ding{55}}
            & \textcolor{darkgreen}{\checkmark} & \textcolor{red}{\ding{55}} & \textcolor{darkgreen}{\checkmark}
            & \textcolor{red}{\ding{55}} & MC+OE \\
%    Perception Test \cite{patraucean2023perception} & 11,620 & 38,060 & – &
%\textcolor{red}{\ding{55}} & \textcolor{darkgreen}{\checkmark} & Vid+Cli+Ins &
%\textcolor{darkgreen}{\checkmark} & \textcolor{darkgreen}{\checkmark} &
%\textcolor{darkgreen}{\checkmark} & \textcolor{red}{\ding{55}} & \textcolor{darkgreen}{\checkmark} &
%\textcolor{red}{\ding{55}} & MC+OE \\
    MVBench \cite{li2024mvbench} & 4,000 & 4,000 & -- &
\textcolor{red}{\ding{55}} & \textcolor{darkgreen}{\checkmark} & Vid+Cli &
\textcolor{darkgreen}{\checkmark} & \textcolor{darkgreen}{\checkmark} &
\textcolor{darkgreen}{\checkmark} & \textcolor{red}{\ding{55}} & \textcolor{darkgreen}{\checkmark} &
\textcolor{red}{\ding{55}} & MC \\
MovieChat-1K \cite{song2024moviechat} & 1,000 & 13,000 & -- &
\textcolor{red}{\ding{55}} & \textcolor{darkgreen}{\checkmark} & Vid+Cli &
\textcolor{red}{\ding{55}} & \textcolor{red}{\ding{55}} &
\textcolor{darkgreen}{\checkmark} & \textcolor{red}{\ding{55}} & \textcolor{red}{\ding{55}} &
\textcolor{red}{\ding{55}} & MC+OE \\

    Video-MME \cite{fu2024video}             & 900  & 2,700 & –   & \textcolor{darkgreen}{\checkmark}        & \textcolor{darkgreen}{\checkmark} & Vid+Cli+Ins
                 & \textcolor{darkgreen}{\checkmark} & \textcolor{red}{\ding{55}}
                 & \textcolor{darkgreen}{\checkmark} & \textcolor{red}{\ding{55}} & \textcolor{red}{\ding{55}}
                 & \textcolor{red}{\ding{55}} & MC \\
    MLVU \cite{zhou2024mlvu}                 & 1,730& 3,102 & –   & \textcolor{darkgreen}{\checkmark}        & \textcolor{darkgreen}{\checkmark} & Vid+Cli+Ins
                 & \textcolor{red}{\ding{55}} & \textcolor{red}{\ding{55}}
                 & \textcolor{red}{\ding{55}} & \textcolor{red}{\ding{55}} & \textcolor{red}{\ding{55}}
                 & \textcolor{red}{\ding{55}} & MC+OE \\
    % -------- Block 3 ----------
    \rowcolor[HTML]{EFEFEF}
    \multicolumn{14}{l}{\textit{Video Text Datasets}} \\
        BovText \cite{wu2021bilingual}          & 2,000   & –     & –   & \textcolor{darkgreen}{\checkmark} & \textcolor{darkgreen}{\checkmark} & Ins
                 & \textcolor{darkgreen}{\checkmark} & \textcolor{darkgreen}{\checkmark}
                 & \textcolor{red}{\ding{55}} & \textcolor{red}{\ding{55}} & \textcolor{red}{\ding{55}}
                 & \textcolor{red}{\ding{55}} & OE \\
    RoadText1k \cite{reddy2020roadtext}      & 1000   & –     & –   & \textcolor{red}{\ding{55}} & \textcolor{red}{\ding{55}}       & Ins
                 & \textcolor{darkgreen}{\checkmark} & \textcolor{darkgreen}{\checkmark}
                 & \textcolor{red}{\ding{55}} & \textcolor{red}{\ding{55}} & \textcolor{red}{\ding{55}}
                 & \textcolor{red}{\ding{55}} & OE \\
    M4ViteVQA \cite{zhao2022towards}         & 680  & 2,103 & 40  & \textcolor{red}{\ding{55}} & \textcolor{darkgreen}{\checkmark} & Cli+Ins
                 & \textcolor{red}{\ding{55}} & \textcolor{red}{\ding{55}}
                 & \textcolor{darkgreen}{\checkmark} & \textcolor{darkgreen}{\checkmark} & \textcolor{red}{\ding{55}}
                 & \textcolor{red}{\ding{55}} & MC+OE \\
    RoadTextVQA \cite{tom2023reading}        & 329  & 1,052 & 60  & \textcolor{red}{\ding{55}} & \textcolor{red}{\ding{55}}       & Cli+Ins
                 & \textcolor{red}{\ding{55}} & \textcolor{red}{\ding{55}}
                 & \textcolor{darkgreen}{\checkmark} & \textcolor{darkgreen}{\checkmark} & \textcolor{red}{\ding{55}}
                 & \textcolor{red}{\ding{55}} & MC \\
 %   LongVideoBench \cite{wu2024longvideobench} & 3,763 & 6,678 & – &
%\textcolor{darkgreen}{\checkmark} & \textcolor{darkgreen}{\checkmark} & Vid+Cli &
%\textcolor{darkgreen}{\checkmark} & \textcolor{red}{\ding{55}} &
%\textcolor{darkgreen}{\checkmark} & \textcolor{red}{\ding{55}} & \textcolor{red}{\ding{55}} &
%\textcolor{red}{\ding{55}} & MC \\

    EgoTextVQA \cite{zhou2025egotextvqa}     & 1,507& 7,064 & 52  & \textcolor{darkgreen}{\checkmark} & \textcolor{red}{\ding{55}}       & Cli+Ins
                 & \textcolor{red}{\ding{55}} & \textcolor{red}{\ding{55}}
                 & \textcolor{darkgreen}{\checkmark} & \textcolor{darkgreen}{\checkmark} & \textcolor{red}{\ding{55}}
                 & \textcolor{red}{\ding{55}} & OE \\
    \rowcolor{ModelGreen}\textbf{Ours}                            & 939  & 2,857 & 65  & \textcolor{darkgreen}{\checkmark} & \textcolor{darkgreen}{\checkmark} & Vid+Cli+Ins
                 & \textcolor{darkgreen}{\checkmark} & \textcolor{darkgreen}{\checkmark}
                 & \textcolor{darkgreen}{\checkmark} & \textcolor{darkgreen}{\checkmark} & \textcolor{darkgreen}{\checkmark}
                 & \textcolor{darkgreen}{\checkmark} & MC+OE \\
    \bottomrule
    \end{tabular}
  \end{adjustbox}

  \label{tab:benchmark-comparison}
  \vspace{-20pt}
\end{table*}

\section{Related Work}

\subsection{Video Large Language Models}
With the rapid advancement of large language models (LLMs), a series of video large language models (Video LLMs) have emerged \cite{liu2023visual,zhu2023minigpt,liu2023llava,chen2024far}, leveraging LLMs as backbones to enhance video reasoning capabilities. Early Video LLMs primarily relied on sparsely sampled frames and temporal modeling mechanisms \cite{li2023blip,liu2024oryx}, such as Q-Former and temporal pooling, to facilitate video captioning and question answering. Building upon these designs, subsequent models \cite{bai2025qwen2,chen2024internvl,zhang2025videollama,li2024videochat,liu2024llavanext,zhang2024long,shu2024video,liu2025video,yuan2025memory} have focused on addressing key challenges in video understanding, including fine-grained semantic alignment, temporal representation, and long-duration video comprehension. For instance, Qwen-VL 2.5 \cite{bai2025qwen2} introduces dynamic resolution processing and absolute temporal encoding to handle variable-resolution videos. Video-LLaMA3 \cite{zhang2025videollama} applies a frame compression strategy based on frame similarity to reduce the number of visual tokens, resulting in more compact and precise video representations. To handle extremely long videos, LongVA \cite{zhang2024long} extends the context length of the LLM backbone and transfers its long-context capability to the video domain. Video-XL \cite{shu2024video} leverages the inherent key-value sparsification mechanism of LLMs to efficiently condense visual inputs. VideoChatFlash \cite{li2024videochat} proposes a hierarchical compression strategy, reducing token redundancy in both the video and language modules.

\subsection{Video Understanding Benchmarks}
With the growing interest in video LLMs, the development of dedicated benchmarks has become increasingly emphasized. Existing benchmarks have been designed for a variety of video understanding tasks, including action reasoning, spatio-temporal inference, video captioning, and long-video comprehension \cite{fu2024video,zhou2024mlvu,li2024mvbench,wu2024longvideobench,xiao2021next,patraucean2023perception,liu2024tempcompass,maaz2023video}.
For example, NeXT-QA \cite{xiao2021next} evaluates temporal reasoning abilities by testing models on the relationships between human actions.
VideoChatGPT-Bench \cite{maaz2023video} focuses on open-ended video conversation, constructing captioning and dialogue tasks to assess generative and interactive capabilities.
TempCompass \cite{liu2024tempcompass} introduces fine-grained temporal perturbations to assess whether models can answer questions based on temporal changes within the video.
To support comprehensive video question answering, MVBench \cite{li2024mvbench} proposes a large-scale benchmark covering 20 distinct subtasks, spanning multiple perception and reasoning dimensions.
For long-video understanding, VideoMME \cite{fu2024video}, MLVU \cite{zhou2024mlvu}, LVBench \cite{wang2024lvbench} and LongVideoBench \cite{wu2024longvideobench} curate diverse and extended-duration videos to evaluate multi-level abilities across extended temporal contexts.

As text carries rich and structured information in videos, several benchmarks have been proposed to evaluate video text understanding \cite{zhao2022towards,wu2021bilingual,reddy2020roadtext,zhou2025egotextvqa,wu2023icdar}, including tasks such as text tracking, spotting, and reasoning.
Specifically, RoadTextVQA \cite{reddy2020roadtext} focuses on autonomous driving scenarios, while EgoTextVQA \cite{zhou2025egotextvqa} targets egocentric perspectives in daily life settings.
In addition, M4-ViteVQA \cite{zhao2022towards} collects videos from nine diverse real-world scenarios, such as shopping, traveling, and movies, to evaluate the generalization capabilities of video-language models.
However, these benchmarks exhibit two notable limitations.
First, their task types are relatively simple, and therefore insufficient for comprehensively evaluating the diverse capabilities of modern video LLMs.
Second, their video categories and language coverage remain limited, often constrained to specific application domains.

\begin{figure*}[t]
    \centering
    \includegraphics[width=0.9\linewidth]{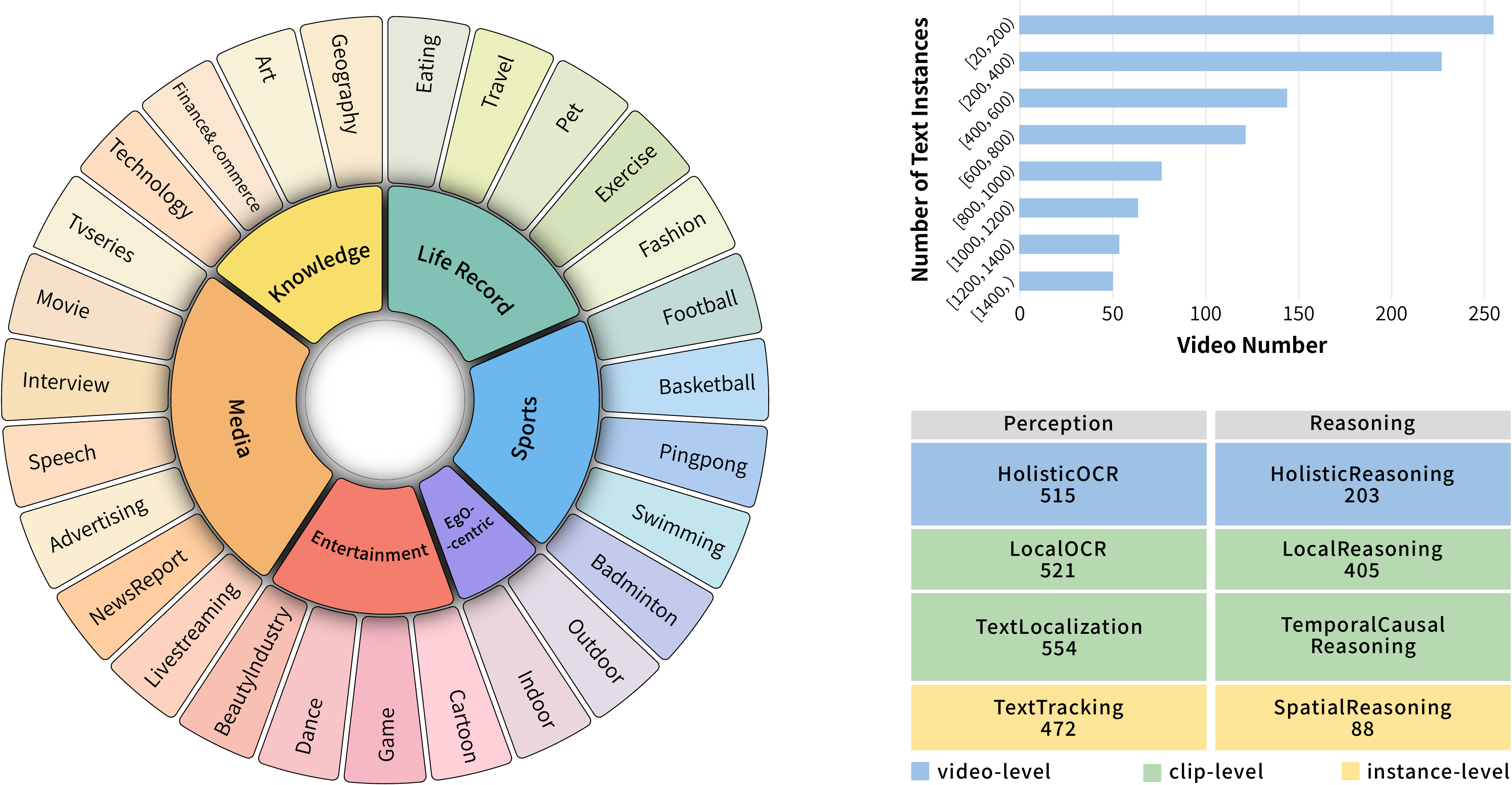}
    \caption{Statistical overview of our VidText. \textbf{(Left)} Video genres included in VidText. \textbf{(Top Right)} Visual Text Instance Distribution. \textbf{(Bottom Right)} Hierarchical Task type settings.}
    \label{fig:statistics}
       \vspace{-15pt}
\end{figure*}

\begin{figure*}[thp]
    \centering
    % \vspace{-0.3cm}
    \includegraphics[width=1.\textwidth]{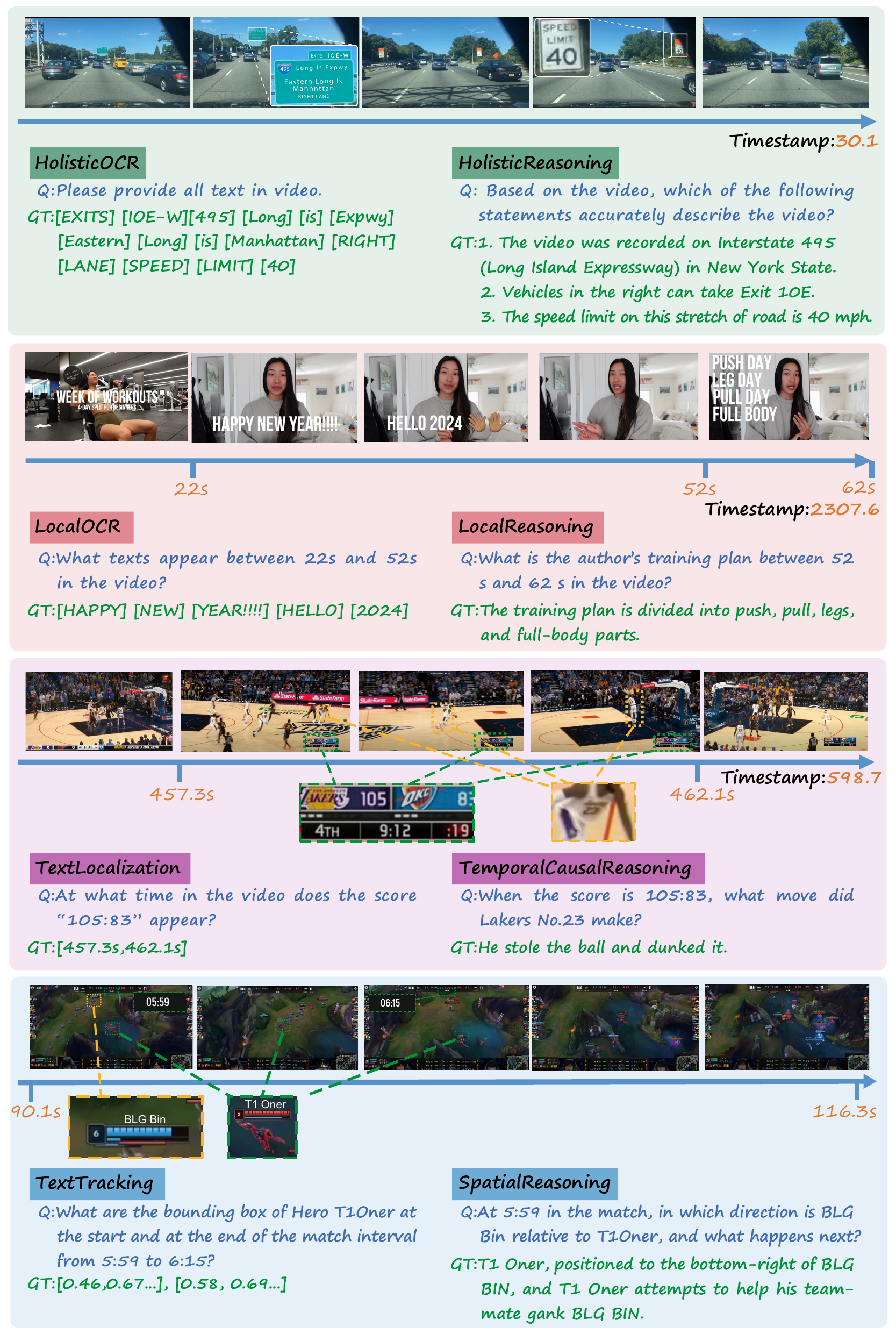}
    % \vspace{-0.3cm}
    \caption{Examples from VidText.
The benchmark includes eight tasks, featuring paired perception and reasoning components designed to evaluate the \textit{video-level}, \textit{clip-level}, and \textit{instance-level} capabilities of LMMs.
Given the video input and textual prompt, models are required to solve the tasks, with ground-truth answers highlighted in green.} 
    \label{fig:pipeline}
    % \vspace{-0.3cm}
\end{figure*}

\section{Dataset Construction}

In this section, we describe the dataset construction process for VidText. We begin by illustrating how the source videos are collected (Sec. \ref{sourcevideo}), followed by a detailed explanation of the annotation pipeline (Sec. \ref{annoge}). Finally, we describe the task taxonomy of our benchmark (Sec. \ref{taskdefination}).

\subsection{Video Collection}
\label{sourcevideo}
In VidText, we aim to evaluate video text understanding across diverse scenarios, including both video category variety and language diversity. While several existing datasets \cite{zhao2022towards,wu2021bilingual,reddy2020roadtext,wu2023icdar} provide detailed text annotations, they all suffer from several key limitations:
(1) Limited scenario diversity: Most datasets focus on specific domains such as indoor scenes or egocentric videos, lacking coverage of richer, more interactive contexts such as sports events, livestreaming games, or daily vlogs.
(2) Lack of language diversity: Nearly all existing datasets contain only English, failing to reflect the multilingual nature of real-world video text.
(3) Short video duration: Many videos are only 10–15 seconds long, which limits their suitability for tasks involving cross-temporal reasoning or holistic understanding.
Therefore, in addition to incorporating existing datasets, we further collect video data from comprehensive long-form video benchmarks \cite{fu2024video,zhou2024mlvu} and public platforms such as YouTube, in order to enhance the scenario diversity, temporal richness, and linguistic coverage of VidText.

For the manually collected videos, we leverage expert models to construct an effective selection pipeline. First, we ensure the presence of visual text in each video by using Gomatching \cite{he2024gomatching}, a video text detection tool, to assess text density. Second, we filter out low-quality videos containing blur, watermarks, or low resolution, using existing video quality assessment models \cite{wen2024modular,mi2024clif}.
Third, we enforce a minimum duration threshold of 3 minutes to guarantee sufficient temporal content. As a result, we collect a total of 939 high-quality videos, each annotated with one of 27 predefined scene categories. Additionally, we record metadata for each video, including language type, resolution, frame rate, and text density. Fig.~\ref{fig:statistics} presents basic statistics of VidText. More detailed statistics across multiple dimensions are provided in the Supplementary Materials.

\subsection{Annotations Generation}
\label{annoge}
To support evaluation at both the perception and reasoning levels, VidText provides meticulously constructed annotations tailored to the requirements of each task.

\textbf{Perception.} For each qualified video, we adopt a bottom-up strategy to construct multi-granularity annotations, including instance-level, clip-level, and video-level information.
First, annotators are instructed to track at least three clear visual text instances throughout the video. For each instance, we conduct frame-by-frame fine-grained annotation until it disappears, generating a sequence of annotations that include bounding boxes, transcriptions, and unique track IDs.
Second, the video is segmented into multiple intervals based on its duration (i.e., longer videos are divided into more segments). For each segment, we check the presence of visual text using instance-level annotations and record clip-level labels, including the temporal span (start and end timestamps) and associated transcriptions.
Third, a separate group of annotators performs video-level annotations, which involve recording all distinct transcriptions that appear across the entire video. Specifically, for Chinese, we use text lines as the basic annotation unit, while for other languages, annotations are performed at the word level.

\textbf{Reasoning.} Since the multi-granularity annotations constructed for perception address the question of ``what texts appear in the video or clip'', we further investigate ``what actions or events are linked to these texts''.
To this end, we design a video text-centric Chain-of-Thought (CoT) annotation pipeline.
First, for each video or clip (as defined by the time span annotations), we apply an adaptive sampling strategy to extract key frames.
Then, we utilize the powerful vision-language model Aria \cite{li2024aria} to generate high-quality frame-level captions, capturing both intra-frame and inter-frame contextual information.
Based on the paired OCR transcripts and the multimodal descriptions, human annotators are instructed to design QA pairs that focus on the semantic or causal relationships between visual text and surrounding visual content.
To ensure the quality of reasoning QA pairs, we enforce two post-validation principles:
(1) Mask the visual texts and verify whether the question can be answered using only the visual content;
(2) Mask the visual frames and check whether the question can be answered using only the textual information.

\subsection{Task Taxonomy}
\label{taskdefination}

Based on the detailed annotations encompassing perception and reasoning, we further define 8 hierarchical tasks which are demonstrated on Fig. \ref{fig:pipeline}.

\textbf{Holistic OCR \& Holistic Reasoning.}
Holistic OCR requires the model to recognize all visual texts appearing throughout the entire video. Redundant entries are removed, and the remaining text instances are sorted in chronological order. We evaluate this task using the F1-score, which is calculated based on instance-level precision and recall.
Holistic Reasoning assesses the model’s ability to understand the overall topic of the video by integrating recognized textual information with global semantic context.  The task is formulated as a multi-label selection problem, where the model is asked to choose three correct answers from seven candidate options. Performance is measured by top-3 accuracy.

\textbf{Local OCR \& Local Reasoning.}
In contrast to holistic tasks, Local OCR and Local Reasoning focus on the model's ability to spot and interpret visual text within user-specified video segments.
Local OCR requires recognizing all visual texts that appear within a given segment and is evaluated using the F1-score based on instance-level matching.
Local Reasoning assesses the model’s ability to infer local semantic meaning or intent from the text. It is formulated as a multiple-choice question, and performance is measured by answer accuracy.

\textbf{Text Localization \& Temporal Causal Reasoning.}
Similar to temporal grounding tasks, Text Localization requires the model to accurately predict the temporal interval during which a specific text appears in the video. The task is evaluated using Mean Intersection-over-Union (mIoU) based on ground-truth temporal spans.
The corresponding reasoning task, Temporal Causal Reasoning, extends beyond localization to assess whether the model can infer causal relationships between identified texts and subsequent multimodal events or actions.
Standard evaluation is conducted in a multiple-choice format, with accuracy as the performance metric.

\textbf{Text Tracking \& Spatial Reasoning.}
Given a target text instance, Text Tracking requires the model to predict its spatial bounding box locations at its first and last appearance within the video.
Spatial Reasoning extends this task by asking the model to infer spatial relationships between the textual instance and surrounding visual elements at a specified timestamp.
To enable standardized evaluation with LMMs, both tasks are formatted as multiple-choice questions.

\begin{table*}[t]
  \centering
 \caption{
  The overall performance on VidText.
  \textbf{HoliOCR}: Holistic OCR;  
  \textbf{HoliRea.}: Holistic Reasoning;  
  \textbf{LocalOCR}: Local OCR;  
  \textbf{LocalRea.}: Local Reasoning;  
  \textbf{TextLocal.}: Text Localization;  
  \textbf{TempCauRea.}: Temporal Causal Reasoning;  
  \textbf{TextTrac.}: Text Tracking;  
  \textbf{SpaRea.}: Spatial Reasoning;  
  \textbf{Avg.}:  the average performance of the eight tasks. The best Accuracy / Score results are highlighted.
}
  \renewcommand{\arraystretch}{1.2}
  \resizebox{\textwidth}{!}{
  \begin{tabular}{l c c c c c c c c c c}
    \toprule
    \textbf{Method} & \textbf{Size} & \textbf{Avg.} &
    \textbf{HoliOCR} & \textbf{HoliRea.} &
    \textbf{LocalOCR} & \textbf{LocalRea.} &
    \textbf{TextLocal.} & \textbf{TempCauRea.} &
    \textbf{TextTrac.} & \textbf{SpaRea.} \\
    \midrule
     \rowcolor[HTML]{EFEFEF} % ← 灰色
    \textit{Human} & -- & 89.5 &
    92.8 & 96.0 & 94.3 & 95.7 & 81.3 & 88.6 & 80.3 & 87.3 \\[-0.2em]
    % -------------------- Closed-source --------------------
    \rowcolor[HTML]{F1F6EC}
    \multicolumn{11}{l}{\textit{\textbf{proprietary  LMMs}}} \\[-0.2em]
    \rowcolor[HTML]{F1F6EC} GPT-4-Turbo \cite{achiam2023gpt}
        & --  & 29.7 & 22.9 & 28.7 & 36.7 & 36.5 & 15.8 & 39.4 & 24.3 & 33.6 \\
    \rowcolor[HTML]{F1F6EC} Gemini 1.5 Flash \cite{team2023gemini}
        & --  & 34.7 & 26.3 & 34.0 & 40.2 & 42.4 & 28.9 & 40.0 & 30.7 & 35.4 \\
    \rowcolor[HTML]{F1F6EC} GPT-4o \cite{hurst2024gpt}
        & --  & 40.2 & 29.5 & 38.9 & 46.0 & 43.3 & 45.5 & 42.5 & 36.2 & 39.8 \\
    \rowcolor[HTML]{F1F6EC} Gemini 1.5 Pro \cite{team2023gemini}
        & --  & \textbf{45.3} & \textbf{34.8} & \textbf{43.6} & \textbf{50.2} & \textbf{50.1} & \textbf{48.7} & \textbf{47.0} & \textbf{40.3} & \textbf{47.9} \\
    \midrule
    % -------------------- Open-source --------------------
    \rowcolor[HTML]{E3F8F8}
    \multicolumn{11}{l}{\textit{\textbf{Open-source LMMs}}} \\[-0.2em]
    \rowcolor[HTML]{E3F8F8} LongVU \cite{shen2024longvu}
        & 3B  & 17.0 &  5.8 & 20.4 & 15.4 & 17.0 & 15.6 & 15.9 & 15.4 & 30.5 \\
    \rowcolor[HTML]{E3F8F8} Qwen2.5-VL \cite{bai2025qwen2}
        & 3B  & 21.1 & 11.4 & 23.2 & 28.5 & 17.8 & 18.7 & 15.4 & 18.3 & 35.3 \\
    \rowcolor[HTML]{E3F8F8} Video-XL-Pro \cite{liu2025video}
        & 3B  & 22.5 & 10.9 & 22.9 & 30.4 & 15.6 & 18.7 & 27.9 & 20.9 & 32.9 \\
    % ---- 6–31 B ----
  
    \rowcolor[HTML]{E3F8F8} LongVA \cite{zhang2024long}
        & 7B  & 19.2 &  4.8 &  5.6 &  3.2 & 46.9 &  4.5 & 28.3 & 29.6 & 30.5 \\
    \rowcolor[HTML]{E3F8F8} MiniCPM-V2.6 \cite{yao2024minicpm}
        & 7B  & 26.5 & 29.2 & 21.2 & 11.4 & 42.9 & 13.3 & 30.3 & 20.5 & 43.2 \\
    \rowcolor[HTML]{E3F8F8} VideoChatFlash \cite{li2024videochat}
        & 7B  & 29.2 & 13.6 & 13.3 &  1.0 & 50.1 & 45.1 & 42.4 & 23.3 & 44.3 \\
    \rowcolor[HTML]{E3F8F8} Qwen2-VL\cite{wang2024qwen2}
        & 7B  & 30.3 & 27.0 & 34.0 & 37.5 & 23.7 & 11.2 & 42.4 & 24.6 & 42.1 \\
    \rowcolor[HTML]{E3F8F8} Qwen2.5-VL \cite{bai2025qwen2}
        & 7B  & 31.9 & 35.9 & 36.0 & 37.0 & 26.5 & 26.5 & 35.4 & 22.4 & 35.2 \\
    \rowcolor[HTML]{E3F8F8} VideoLLaMA3 \cite{zhang2025videollama}
        & 7B  & \textbf{39.9} & 23.5 & 31.5 & 39.2 & 41.2 & \textbf{47.3} & \textbf{55.6} & 31.1 & 50.0 \\
    \rowcolor[HTML]{E3F8F8} ShareGPT4Video \cite{chen2024sharegpt4video}
        & 8B  & 16.4 &  2.5 &  2.6 &  0.8 & 43.5 &  0.0 & 27.3 & 28.0 & 26.1 \\
    % ---- 32–70 B ----
    \rowcolor[HTML]{E3F8F8} Oryx-1.5 \cite{liu2024oryx}
        & 32B & 35.4 & 35.3 & 33.9 & 30.8 & 48.5 & 26.7 & 45.2 & 26.0 & 36.4 \\
    % ---- ≥71 B ----
    \rowcolor[HTML]{E3F8F8} LLava-OV\cite{li2024llava}
        & 72B & 36.1 & 20.1 & 28.1 & \textbf{41.3} & 49.4 &  9.9 & 54.6 & \textbf{31.8} & \textbf{53.4} \\
    \rowcolor[HTML]{E3F8F8} Qwen2.5-VL \cite{bai2025qwen2}
        & 72B & 38.5 & 40.1 & \textbf{49.3} & 35.9 & 28.2 & 28.7 & 52.5 & 31.1 & 42.1 \\
    \rowcolor[HTML]{E3F8F8} InternVL2.5 \cite{chen2024expanding}
        & 78B & 39.8 & \textbf{40.2} & 37.4 & 29.0 & \textbf{50.4} & 30.5 & 48.5 & 29.9 & 52.3 \\
    \bottomrule
  \end{tabular}}
  \label{tab:main_result}
     \vspace{-20pt}
\end{table*}

\section{Experiments}

\subsection{Settings}

We conduct a comprehensive evaluation of 18 large multimodal models (LMMs) using our VidText benchmark, encompassing both open-source and proprietary models. For proprietary models, we evaluate the Gemini series \cite{team2023gemini} and GPT series \cite{ achiam2023gpt,hurst2024gpt}, using their official multi-image evaluation APIs.
For open-source models, we select current state-of-the-art video LMMs with diverse architectures and LLM sizes, enabling a broad assessment of video text understanding capabilities.
All evaluations are conducted in a zero-shot manner. More details about the evaluation settings are provided in the Supplementary Materials.

\subsection{Main Results}

The overall evaluation results for all investigated LMMs in the VidText are shown in Tab. \ref{tab:main_result}. Individual performances are reported
for each task, while average performances are provided. From the results, we derive three primary conclusions:

1) \textbf{Gemini 1.5 Pro \cite{team2023gemini} achieves the best performance on our benchmark}. It significantly outperforms other models on video-text-based perception and reasoning tasks.

2) \textbf{Proprietary models typically perform better than open-source models}. However, some open-source models deliver surprisingly strong results on specific tasks. For example, VideoLLaMA3 \cite{zhang2025videollama} achieves the highest performance on both Temporal Causal Reasoning and Spatial Reasoning.

3) \textbf{Video text understanding remains an overwhelming challenge for current video LMMs}.
First, even the best models still fall far short of human-level performance.
Second, most LMMs show limited ability in fundamental video OCR tasks, where specialized video OCR models often perform better.
Third, multimodal reasoning based on visual text cues in videos is significantly more difficult than in images: all video multiple-choice reasoning tasks yield accuracies below 60\%, falling far behind the performance on similar image-based tasks. (ST-VQA \cite{biten2019scene} and Text-VQA \cite{singh2019towards})

Beyond the primary conclusions on overall performance, we further analyze model behaviors across individual tasks.

4) \textbf{Among multi-granular tasks, video-level and instance-level tasks are more challenging than clip-level tasks, across both perception and reasoning settings.}
We hypothesize that this is due to the limited capabilities of current LMMs in two aspects: video-level tasks require global information aggregation, while instance-level tasks demand fine-grained retrieval and grounding, both of which remain weak points for existing models.

5) \textbf{For video-level and instance-level tasks, the performance of perception and reasoning shows a strong correlation, while the two appear relatively independent in clip-level tasks.}
This may be because certain clip-level perception tasks, such as text localization, require accurate temporal grounding based on fine-grained visual cues.
However, the corresponding reasoning tasks, such as temporal reasoning, can often be solved using local visual clues from sparsely sampled frames, allowing models to bypass the need for precise perception outputs.

6) \textbf{Scaling up the size of LLMs leads to more significant performance gains on reasoning tasks compared to perception tasks.}
This suggests that video text perception cannot be effectively improved by model scale alone, and instead requires careful architectural design, specialized training data, and other task-specific considerations.

%%%%%%%%%%%%%%%%%%%%%%%%%%%%%%%%%%%%%%%%%%%%%%%%%%%%%%%%%%%%%%%%%%%%%%%%
\section{Ablation Studies}
\label{sec:ablation}

This section begins with an investigation into the effectiveness of our hierarchical task design, including the multi-granularity structure and the extension from perception to reasoning.
We also explore critical factors that affect model performance in video text understanding through a series of ablation studies.

\subsection{Investigating the effectiveness of VidText Design}

\begin{figure*}[t]
    \centering
    \includegraphics[width=0.9\linewidth]{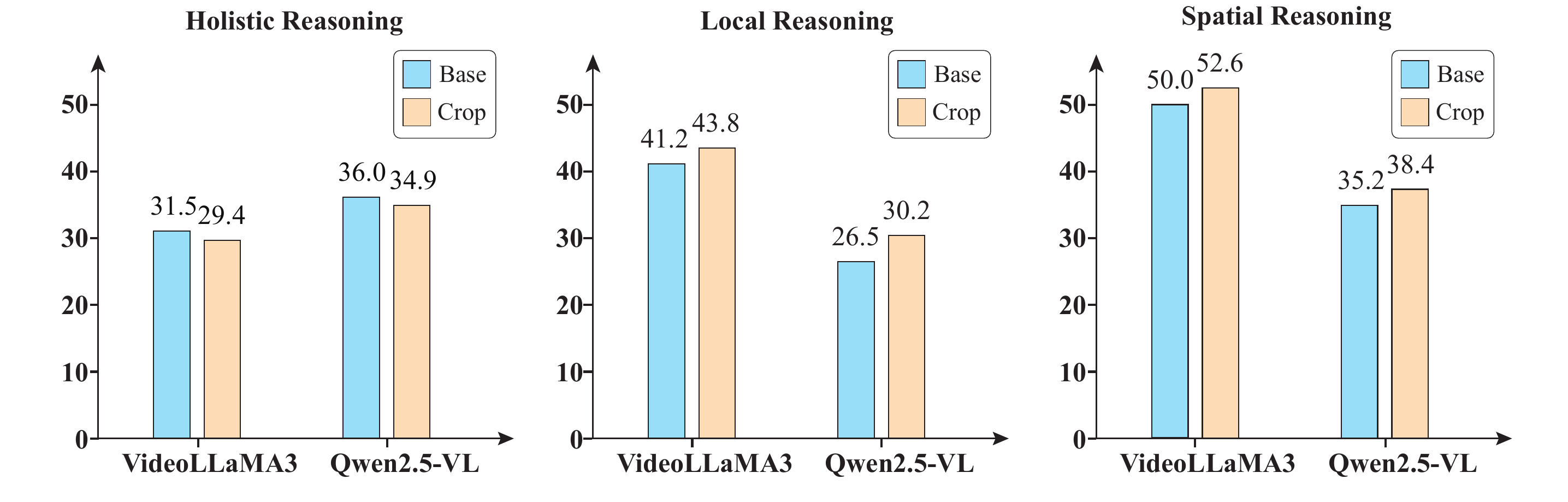}
    \caption{Ablation studies on the multi-granularity design of VidText.}
    \label{fig:multi_granu}
       \vspace{-5pt}
\end{figure*}

\textbf{Multi-granularity design.}
VidText includes multi-granular tasks spanning video-level, clip-level, and instance-level. To verify that tasks at different levels require correspondingly different levels of contextual information, we conduct ablation studies using VideoLLaMA3 \cite{zhang2025videollama} and Qwen2.5-VL \cite{bai2025qwen2}.
Specifically, for holistic tasks, we randomly extract 50\% of the video duration as a segment and evaluate performance on Holistic Reasoning. For clip-level and instance-level tasks, we select key clips based on their original task annotations.
As shown in Fig. \ref{fig:multi_granu}, clip-level and instance-level tasks benefit significantly from segment-based evaluation, as key frames provide concentrated visual text information.
In contrast, Holistic Reasoning performance declines, as the task requires global information aggregation, which is lost when only partial segments are used.

\begin{figure*}[t]
    \centering
    \includegraphics[width=0.9\linewidth]{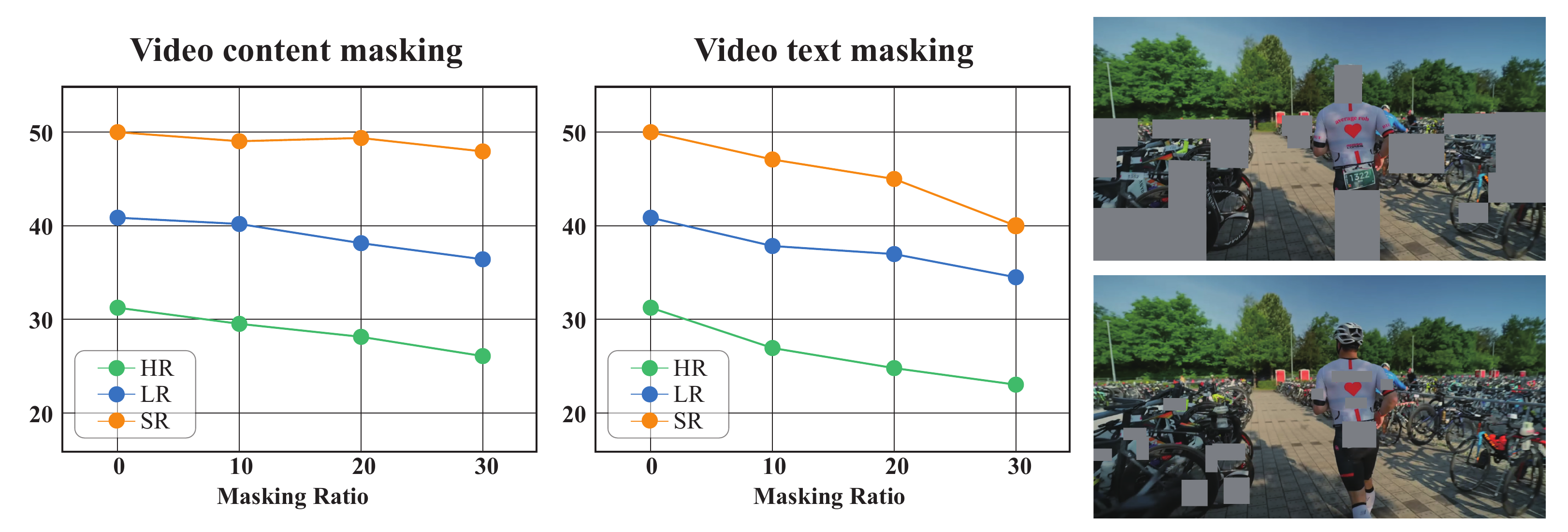}
    \caption{Ablation studies on the joint reasoning of video texts and video contents. ``HR'', ``LR'' and ``SR'' denote Holistic Reasoning, Local Reasoning and Spatial Reasoning, respectively. We visualize ``Video content masking'' and ``Video Text masking'' in the right part.}
    \label{fig:masking}
    \vspace{-5pt}
\end{figure*}

\textbf{Joint video text and multimodal contexts reasoning.}
VidText successfully extends perception-level tasks into reasoning tasks, which require the joint modeling of video texts and their multimodal contextual information.
To validate this, we perform an ablation study by selectively masking either the visual text regions or the surrounding video content at varying random ratios.
As shown in Fig. \ref{fig:masking}, the performance on all reasoning tasks consistently drops as the masking ratio increases, confirming that both textual and visual cues are essential for reasoning under our task design.

% \begin{table*}[tp]
%   \centering
%   \small
%   \setlength\tabcolsep{5pt}
%   \begin{tabular*}{0.75\textwidth}{@{\extracolsep{\fill}} lcccc}
%     \toprule
%     \multicolumn{5}{c}{\textbf{A. Impact of Video Granularity (Full vs.\ Local Clip)}} \\
%     \midrule
%     \multirow{2}{*}{\textbf{Model}} &
%     \multicolumn{2}{c}{HOCR (F1)$\uparrow$} &
%     \multicolumn{2}{c}{TCR (Acc.)$\uparrow$} \\
%     \cmidrule(lr){2-3} \cmidrule(lr){4-5}
%     & Full & Local & Full & Local \\
%     \midrule
%     VideoLLaMA3     & \textbf{20.3} & 17.7\textcolor{red}{\,($\downarrow$2.6)}   & 43.3 & \textbf{56.7}\textcolor{darkgreen}{\,($\uparrow$13.4)} \\
%     Qwen2.5-VL-7B   & \textbf{36.3} & 35.3\textcolor{red}{\,($\downarrow$1.0)}   & 26.7 & \textbf{33.3}\textcolor{darkgreen}{\,($\uparrow$6.6)}  \\
%     \bottomrule
%   \end{tabular*}
%   \caption{Full video vs.\ local clip on Holistic OCR (HOCR) and Temporal–Causal Reasoning (TCR).}
%   \label{tab:granularity}
% \end{table*}

\begin{table}[tp]
    \centering
    \setlength\tabcolsep{2.5pt}
    \resizebox{1.0\linewidth}{!}{
        \begin{tabular}{l c l||l c l||l c l}
        \Xhline{1.0pt}
        \multicolumn{3}{c||}{\textbf{Impact of input resolution}} & \multicolumn{3}{c||}{\textbf{Impact of OCR ability}} &   \multicolumn{3}{c}{\textbf{Impact of LLM}} \\
        \Xhline{0.8pt}
        Model & resolution &Avg & Model & OCRBench &Avg & Model & LLM & Avg\\
        \Xhline{0.8pt}
        \multirow{2}{*}{Oryx-1.5} & $448^2$ & 35.4 & LLaVA-OV & 621 &36.1 & \multirow{2}{*}{LLaVA-OV} &Qwen2-7B &22.1 \\
        % \hline
        ~ & $896^2$ & 38.6\darkGreen{$\uparrow$3.2} & VideoLLaMA3 & 828 &39.9\darkGreen{$\uparrow$3.8}  &~ & Qwen2-72B & 36.1\darkGreen{$\uparrow$14.0}    \\
        \hline
        \multirow{2}{*}{InternVL} & $448^2$ & 39.8 & GPT4V& 645 & 29.7 & \multirow{2}{*}{LLaVA-Next} & LLaMA3-8B & 15.3 \\
        % \hline
        ~ & $896^2$ & 44.8\darkGreen{$\uparrow$5.0}  & GPT-4o & 822&40.2\darkGreen{$\uparrow$10.5}  & ~ & Qwen2-7B &20.8 \darkGreen{$\uparrow$5.2}  \\
        \Xhline{1.0pt}
        \end{tabular}
    }
    \vspace{0.2cm}
    \caption{Detailed analysis about the impact of input resolution, image OCR ability, and LLM Backbone. LLaVA-Next: LLaVA-Next-Video \cite{liu2024llavanext}.} 
    \label{tab:different_impact}
    \vspace{-0.5cm}
\end{table}

\subsection{Exploring Crucial Factors of Video Text Understanding }

\textbf{Model-intrinsic Factors.}
As shown in Tab. \ref{tab:different_impact}, we conduct ablation studies on several influential factors.
First, we examine the \textit{impact of input resolution} using two representative models, Oryx-1.5 \cite{liu2024oryx} and InternVL2.5 \cite{chen2024internvl}, both of which support adjustable input sizes. Increasing the resolution significantly improves video text understanding performance, especially in InternVL2.5 \cite{chen2024internvl}, where the input images are divided into sub-patches, where higher resolution allows better preservation of text details.
Second, to assess the \textit{role of OCR capability}, we refer to each model’s performance on standard OCR benchmarks such as OCRBench \cite{liu2024ocrbench}. The results show that a model’s video text understanding performance generally aligns with its fundamental OCR accuracy.
Finally, we compare 
\textit{different LLM backbones} and find that certain architectures (e.g., Qwen2.5) exhibit stronger performance in multilingual scenarios, often outperforming LLaMA-based variants.
These observations collectively indicate that video text understanding is influenced by a combination of input fidelity, OCR capacity, and language modeling strength.

\textbf{External Factors.} As demonstrated in Tab. \ref{tab:ab_external}, we first investigate whether external auxiliary information can enhance video text understanding, particularly for reasoning tasks. In this study, we consider audio transcripts and video text (e.g., subtitles or OCR outputs), both of which can be extracted using specialized tools. We convert these modalities into textual sequences and append them to the original query as contextual subtitles.
As shown in our experiments, both sources contribute positively to performance. Video text provides stronger gains in global tasks that require long-range context, while audio transcripts are more beneficial for local tasks, possibly due to their alignment with short-term actions or events. Second, we propose a video text-centric Chain-of-Thought (CoT) reasoning strategy, which decomposes complex reasoning processes into structured sub-steps. Specifically, the video is uniformly segmented into multiple clips. For each clip, the model is prompted to:
(1) spot all visible texts,
(2) generate a detailed description of the clip, and
(3) infer whether any visual texts are semantically related to the description and answer the reasoning question accordingly. This CoT-based prompting strategy yields consistent improvements across all reasoning tasks, highlighting the potential of test-time reasoning augmentation for video-language models.

\begin{table*}[t]
    \centering
    \vspace{5pt}
    \renewcommand{\arraystretch}{1.1}
    \begin{minipage}{0.53\textwidth}
    \centering
    \begin{adjustbox}{max width=\textwidth}
    \begin{tabular}{l cccc}
    \toprule
    \textbf{Method} & \textbf{HR} & \textbf{LR}& \textbf{TR}& \textbf{SR} \\
    \midrule
    Qwen2.5-VL               & 36.0 & 26.5 & 35.4    &35.2     \\
    Qwen2.5-VL + Audio & 36.3 & 26.6 & 35.2    &35.4     \\
    Qwen2.5-VL + Text & 37.2 & 28.3 &37.9     &38.1     \\
    Qwen2.5-VL + Audio + Text  & \textbf{37.6} & \textbf{29.5} & \textbf{38.0}    &\textbf{39.5}     \\
    \bottomrule
    \end{tabular}
    \end{adjustbox}
    \end{minipage}
    \hfill
    \begin{minipage}{0.45\textwidth}
    \centering
    \begin{adjustbox}{max width=\textwidth}
    \begin{tabular}{l cccc}
    \toprule
    \textbf{Method} & \textbf{HR} & \textbf{LR}& \textbf{TR}& \textbf{SR} \\
    \midrule
    Qwen2.5-VL           &36.0 &26.5 &35.4 &35.2  \\
    Qwen2.5-VL + CoT     &\textbf{40.5} &\textbf{28.7} &\textbf{37.2} &\textbf{40.9}  \\
    \midrule
    VideoLLaMA3          &31.5 &41.2 &55.6 &50.0  \\
    VideoLLaMA3 + CoT    &\textbf{33.8} &\textbf{44.6} &\textbf{56.2} &\textbf{53.8} \\
    \bottomrule
    \end{tabular}
    \end{adjustbox}
    \end{minipage}
    \caption{Ablations about auxiliary information \textbf{(Left)} and CoT strategy \textbf{(Right)} to video text understanding.``HR'', ``LR''. ``TR'' and ``SR'' mean Holistic Reasoning, Local Reasoning, Temporal Causal Reasoning and Spatial Reasoning. }
    \label{tab:ab_external}
    % \vspace{-0.5cm}
\end{table*}

\section{Conclusion}
%%%%%%%%%%%%%%%%%%%%%%%%%%%%%%%%%%%%%%%%%%%%%%%%%%%%%%%%%%%%
% ==================  Table A: Granularity  =====================
This paper presents VidText, a novel benchmark for evaluating video text understanding in large multimodal models (LMMs). With three key innovations, including broad scenario and multilingual coverage, a multi-granular evaluation framework, and paired perception–reasoning tasks, VidText enables comprehensive and in-depth analysis of LMM performance on video text understanding. Empirical studies reveal that current LMMs still face significant challenges in both perceiving and reasoning over video texts.
Future progress in this area will require the joint optimization of multiple complex factors, including model-intrinsic aspects (such as input resolution, OCR capability, and LLM backbone) and external strategies (such as auxiliary modality integration and Chain-of-Thought prompting). We hope VidText will serve as a valuable resource for advancing research in the OCR and video understanding communities.

\clearpage
\newpage
\bibliography{main}{}
\bibliographystyle{unsrt}
%%%%%%%%%%%%%%%%%%%%%%%%%%%%%%%%%%%%%%%%%%%%%%%%%%%%%%%%%%%%
\clearpage
\appendix

\section{Overview of Appendix}

\begin{itemize}
\item \ref{appendix:limitation}: \textbf{Limitations}.
    \item  \ref{appen:broad}: 
        \textbf{Broader Impact}.
    \item  \ref{appendix:Collecting Details of VidText}: 
    \textbf{Collecting Details of VidText}.
    \item  \ref{appendix:annotation}: \textbf{Details of Annotation}.
    \item \ref{appendix:settings}: \textbf{Detailed Experimental Results}.
    \item \ref{appendix:prompts}: \textbf{Model Prompt}.
     \item \ref{appendix:vis}: \textbf{More Visualization Results}.
    \item \textbf{Checklist} 
\end{itemize}

\section{Limitation}
\label{appendix:limitation}

%%%%%%%%%%%%%%%%%%%%%%%%%%%%%%%%%%%%%%%%%%%%%%%%%%%%%%%%%%%%%%%%%%%%%%%%
% Appendix  B  —  Limitations
%%%%%%%%%%%%%%%%%%%%%%%%%%%%%%%%%%%%%%%%%%%%%%%%%%%%%%%%%%%%%%%%%%%%%%%%
We summarize the limitations of our work as follows:
\begin{itemize}[leftmargin=1.2em]
\item \textbf{Limited scenario coverage:}  
  Although VidText includes 27 fine-grained video categories, it still lacks representation of long-tail or high-risk domains such as medical emergencies, industrial workflows, or disaster scenarios.

  \item \textbf{Imbalanced language distribution:}  
  The majority of samples are in English and Chinese, with significantly fewer examples in other languages such as German, Korean, and Japanese. This imbalance prevents a thorough evaluation of multilingual OCR and reasoning capabilities.

  \item \textbf{Scarcity of challenging text instances:}  
  VidText contains relatively few examples involving difficult text conditions such as severe occlusion, low resolution, motion blur, unusual fonts, or multi-line arrangements. This limits the benchmark’s ability to fully assess model robustness under real-world noise and distortion.
\end{itemize}

% We discuss the limitations of our work from two perspectives: dataset-level constraints inherent to VidText and model-level generalization bottlenecks. These two aspects are interconnected and jointly shape the current challenges in video text understanding.

% \subsection{Dataset-Level Limitations}

% \begin{itemize}[leftmargin=1.2em]
%   \item \textbf{Limited scenario coverage:}  
%   Although VidText includes 27 fine-grained video categories, it still lacks representation of long-tail or high-risk domains such as medical emergencies, industrial workflows, or disaster scenarios.

%   \item \textbf{Imbalanced language distribution:}  
%   The majority of samples are in English and Chinese, with significantly fewer examples in other languages such as German, Korean, and Japanese. This imbalance prevents a thorough evaluation of multilingual OCR and reasoning capabilities.

%   \item \textbf{Scarcity of challenging text instances:}  
%   VidText contains relatively few examples involving difficult text conditions such as severe occlusion, low resolution, motion blur, unusual fonts, or multi-line arrangements. This limits the benchmark’s ability to fully assess model robustness under real-world noise and distortion.
% \end{itemize}

% \subsection{Model-Level Generalization Constraints}

\subsection{Discussion}

These dataset and model limitations are mutually reinforcing. Dataset gaps may conceal important weaknesses in current models, while existing models’ deficiencies highlight the need for broader and more diverse benchmarks. Future efforts should focus on expanding long-tail scene and language coverage in VidText, while also improving LMM architectures with better multilingual OCR, noise robustness, and cross-modal reasoning abilities. Furthermore, we also summarize three insights as follows:

\begin{itemize}[leftmargin=1.2em]
  \item \textbf{Weak cross-domain transfer:}  
  Most LMMs are pretrained on image-based OCR tasks and struggle to generalize to unseen video scenes, such as sports broadcasts or livestream interfaces, where text appearance and context are highly dynamic.

  \item \textbf{Insufficient multilingual alignment:}  
  Current models show limited ability in detecting, transcribing, and semantically linking non-English texts to the visual context, resulting in degraded performance on multilingual content.

  \item \textbf{Low robustness to visual noise:}  
  Models often fail when confronted with noisy, blurry, or occluded text, particularly in tasks requiring instance-level grounding. This degrades downstream reasoning performance and reflects a need for stronger visual resilience.
\end{itemize}

\section{Broader Impact}
\label{appen:broad}
The VidText benchmark is poised to make a significant contribution to both the OCR and video understanding communities by bridging the gap between low-level text perception \cite{shu2025visual,li2024first,zeng2024textctrl} and high-level semantic reasoning \cite{long2021scene,zhu2016scene} in video contexts.

For the OCR community, VidText offers a valuable opportunity to move beyond traditional image-based text detection and recognition \cite{huang2022swintextspotter,zhou2017east,liao2020real,shu2023perceiving}. By shifting the focus to temporal and contextual dynamics in videos, it promotes the development of algorithms that can track, ground, and interpret visual texts over time.

For the video understanding community, VidText introduces the underexplored yet semantically rich modality of scene text into the landscape of video-language research. By incorporating fine-grained text perception tasks and their paired reasoning counterparts, VidText pushes video-language models to integrate visual texts with multimodal contextual cues, fostering more explainable, interpretable, and grounded video understanding. 

%-------------------------------------------------------------
%%%%%%%%%%%%%%%%%%%%%%%%%%%%%%%%%%%%%%%%%%%%%%%%%%%%%%%%%%%%%%%%%%%%%%%%
% C. Collecting Details of VidText
%%%%%%%%%%%%%%%%%%%%%%%%%%%%%%%%%%%%%%%%%%%%%%%%%%%%%%%%%%%%%%%%%%%%%%%%
\section{Collecting Details of \textsc{VidText}}
\label{appendix:Collecting Details of VidText}

This section outlines the procedures for sourcing, filtering, and analyzing the video content in \textsc{VidText}.

\paragraph{Sources.}  
To ensure a broad coverage of video scenarios and textual styles, \textsc{VidText} integrates data from six public datasets:
\begin{itemize}[leftmargin=1.5em]
    \item \textbf{BOVText}~\cite{wu2021bilingual} — Multi-scene videos suitable for holistic OCR tasks.
    \item \textbf{RoadText-1K}~\cite{reddy2020roadtext} — Dense road-text detection in driving scenarios.
    \item \textbf{DSText}~\cite{wu2023icdar} — Subtitles from indoor instructional videos.
    \item \textbf{M4-ViteVQA}~\cite{zhao2022towards} — Clip- and instance-level multimodal QA videos.
    \item \textbf{Video-MME/MLVU}~\cite{fu2024video,zhou2024mlvu} — Long-form videos with strong temporal reasoning demands.
\end{itemize}

\paragraph{YouTube Supplementation.}  
To supplement long-form data, we collect additional videos from YouTube, focusing on the following categories:
\begin{itemize}[leftmargin=1.5em]
    \item \textbf{Sports highlights:} NBA, FIFA World Cup, and related competitions.
    \item \textbf{Gaming commentary:} live streams and post-game analysis.
    \item \textbf{TV shows and variety entertainment.}
\end{itemize}

\paragraph{Retrieval and Filtering Criteria.}
Candidate videos were retrieved using targeted keyword queries such as \textit{"match subtitles"}, \textit{"game commentary"}, and \textit{"captioned recap"}. We applied the following filtering rules:
\begin{itemize}[leftmargin=1.5em]
    \item \textbf{Minimum duration:} $\geq$3 minutes for YouTube, $>$30 minutes for Video-MME.
    \item \textbf{Scene-text richness:} We use the latest detector \textbf{Gomatching}~\cite{he2024gomatching} to calculate the proportion of frames containing text.
    \item \textbf{Density thresholds:} Videos must meet a minimum ratio of text-bearing frames: \textbf{20\%} for YouTube videos and \textbf{10\%} for Video-MME.
\end{itemize}

\paragraph{Metadata Statistics.}  
We also collect metadata such as video length, resolution, and frame rate to ensure coverage diversity across temporal and visual characteristics.

\begin{figure}[t]
  \centering
  \includegraphics[width=0.8\linewidth]{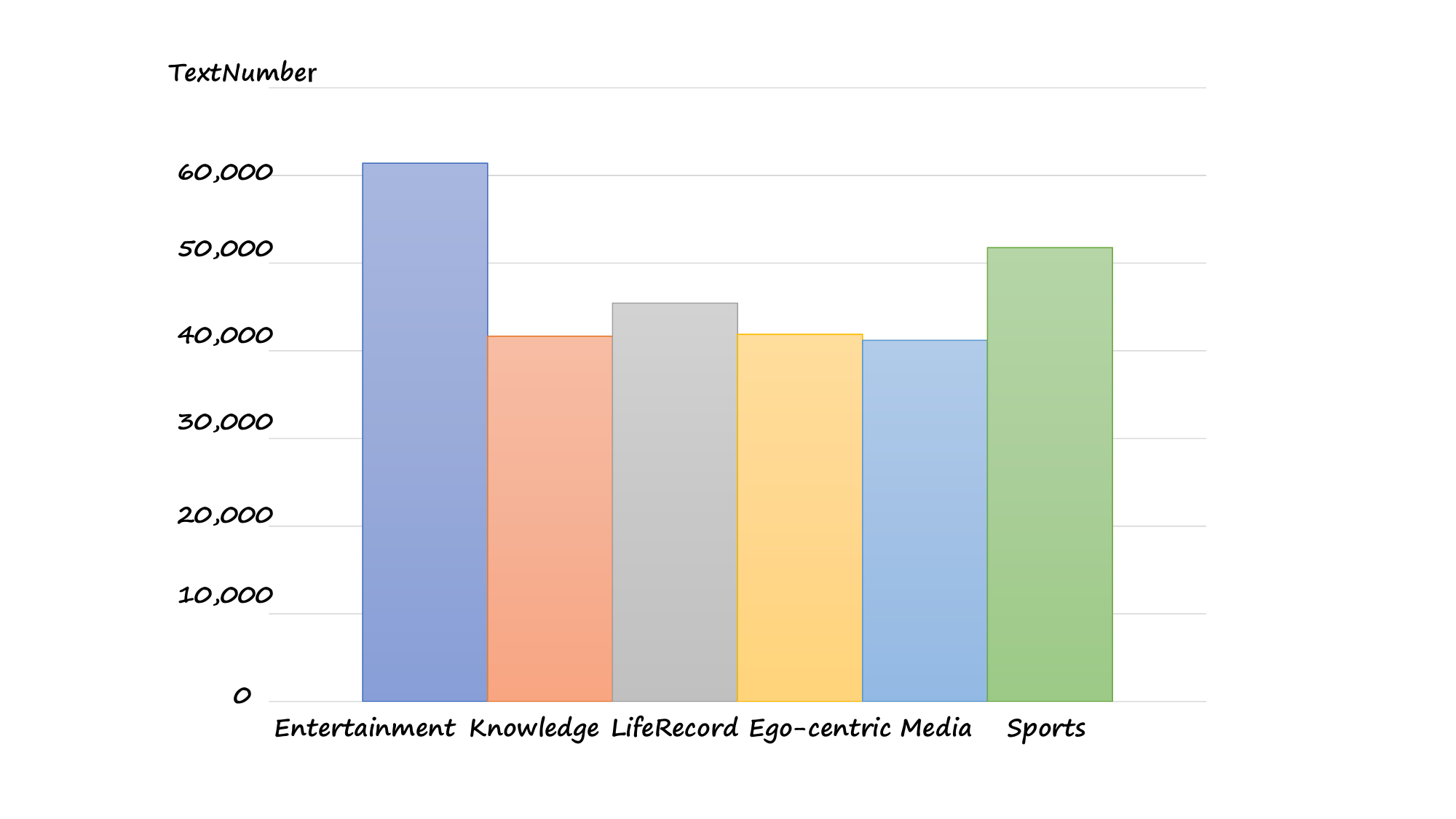}
  \caption{Text quantity distribution across six scene categories.}
  \label{fig:CategoryTextNum}
\end{figure}

\paragraph{Scene and Language Distributions.}  
Fig.~\ref{fig:CategoryTextNum} illustrates the distribution of visual text quantity across six video scene categories. The largest number of text instances appears in \textbf{Entertainment} and \textbf{Sports}-related content, while \textbf{knowledge} and \textbf{Media} are less dense in text content.

\begin{figure}[t]
  \centering
  \includegraphics[width=0.8\linewidth]{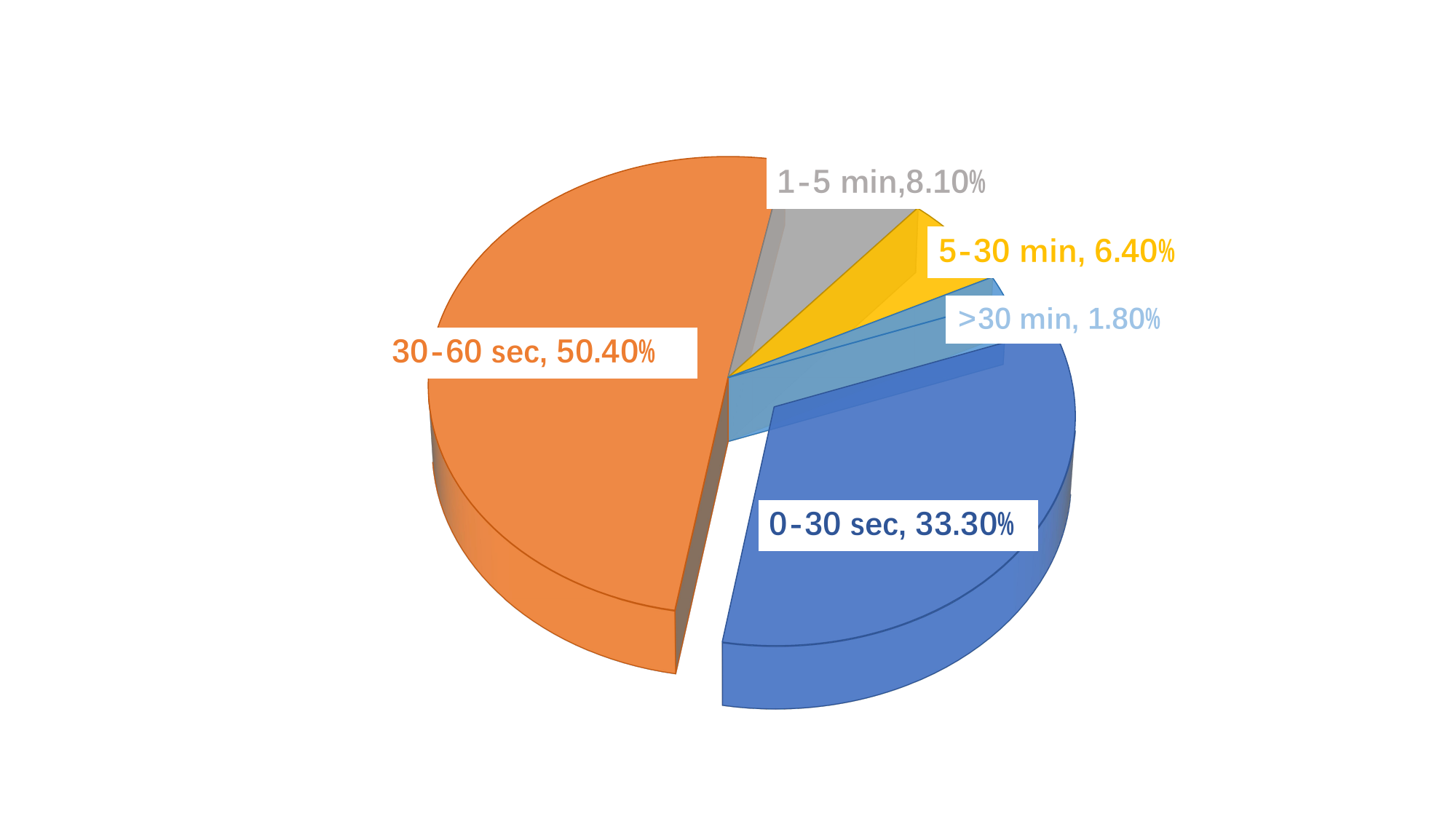}
  \caption{Video duration distribution in \textsc{VidText}.}
  \label{fig:duration}
\end{figure}

\paragraph{Video Duration Distribution.}  
\textsc{VidText} exhibits a wide range of video durations, with an average length of 108.2 seconds. As shown in Fig.~\ref{fig:duration}, this highlights the multi-duration characteristic of \textsc{VidText}, ensuring the temporal diversity needed to support both short-form and long-form video understanding tasks.

\begin{figure}[t]
  \centering
  \includegraphics[width=0.6\linewidth]{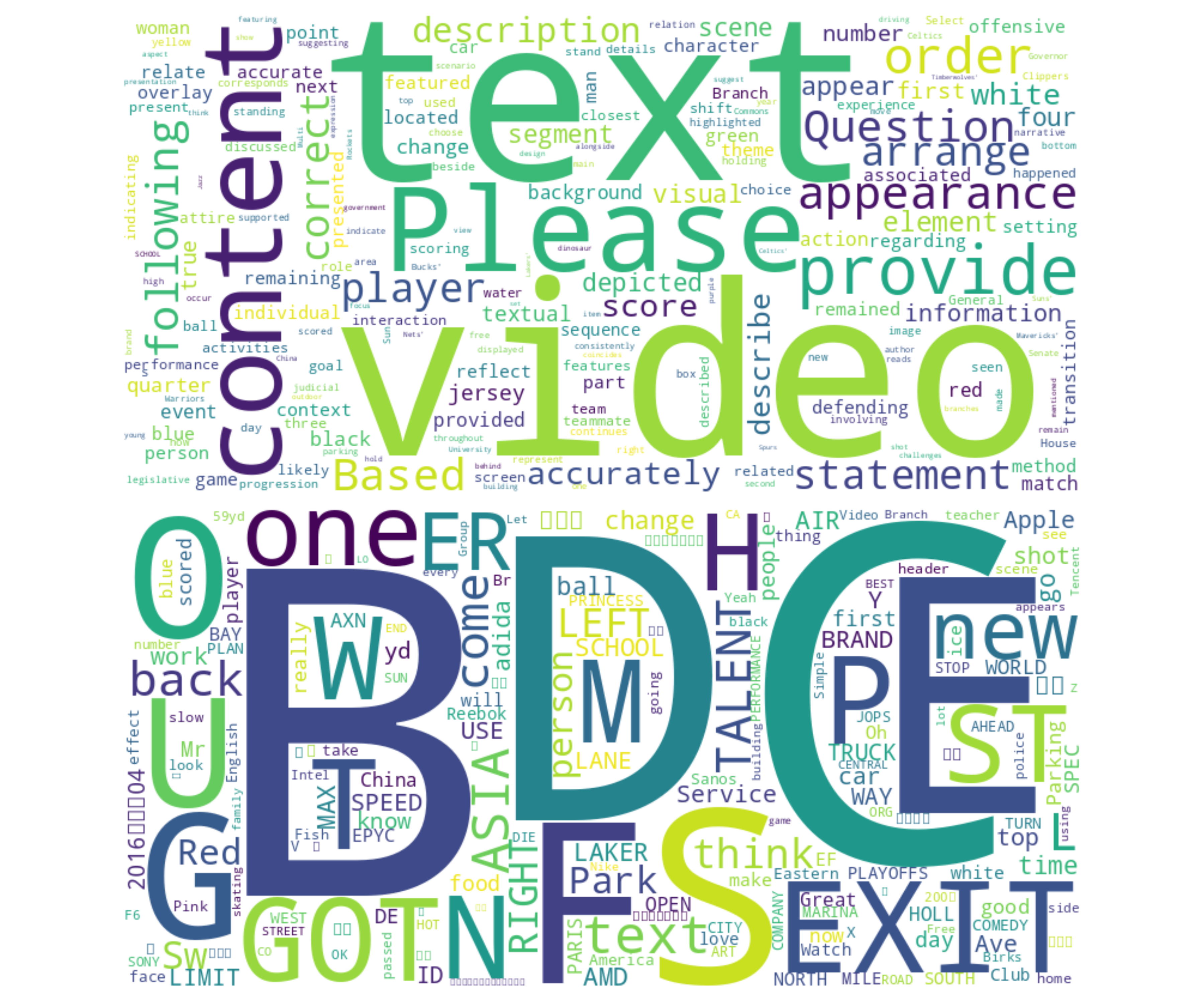}
  \caption{Word cloud of all questions and answers in \textsc{VidText.}}
  \label{fig:vidtext_wordcloud}
\end{figure}

\paragraph{Semantic Content Word Cloud.}  
To visualize the semantic richness and diversity of video-text interactions, we construct a word cloud using all questions and answers in \textsc{VidText}. As shown in Fig.~\ref{fig:vidtext_wordcloud}, high-frequency words such as \textit{text}, \textit{video}, \textit{content}, and \textit{EXIT} reflect a strong alignment between text and semantic reasoning. The co-existence of spatial keywords (e.g., \textit{LEFT}, \textit{RIGHT}), functional terms (e.g., \textit{score}, \textit{speed}), and contextual references (e.g., \textit{player}, \textit{talent}) highlights the multi-granular reasoning needs of the dataset.

\section{Details of Annotation}
\label{appendix:annotation}
\subsection*{D.1 Instance Annotation}
\begin{figure*}[h]
    \centering
    \includegraphics[width=1\textwidth]{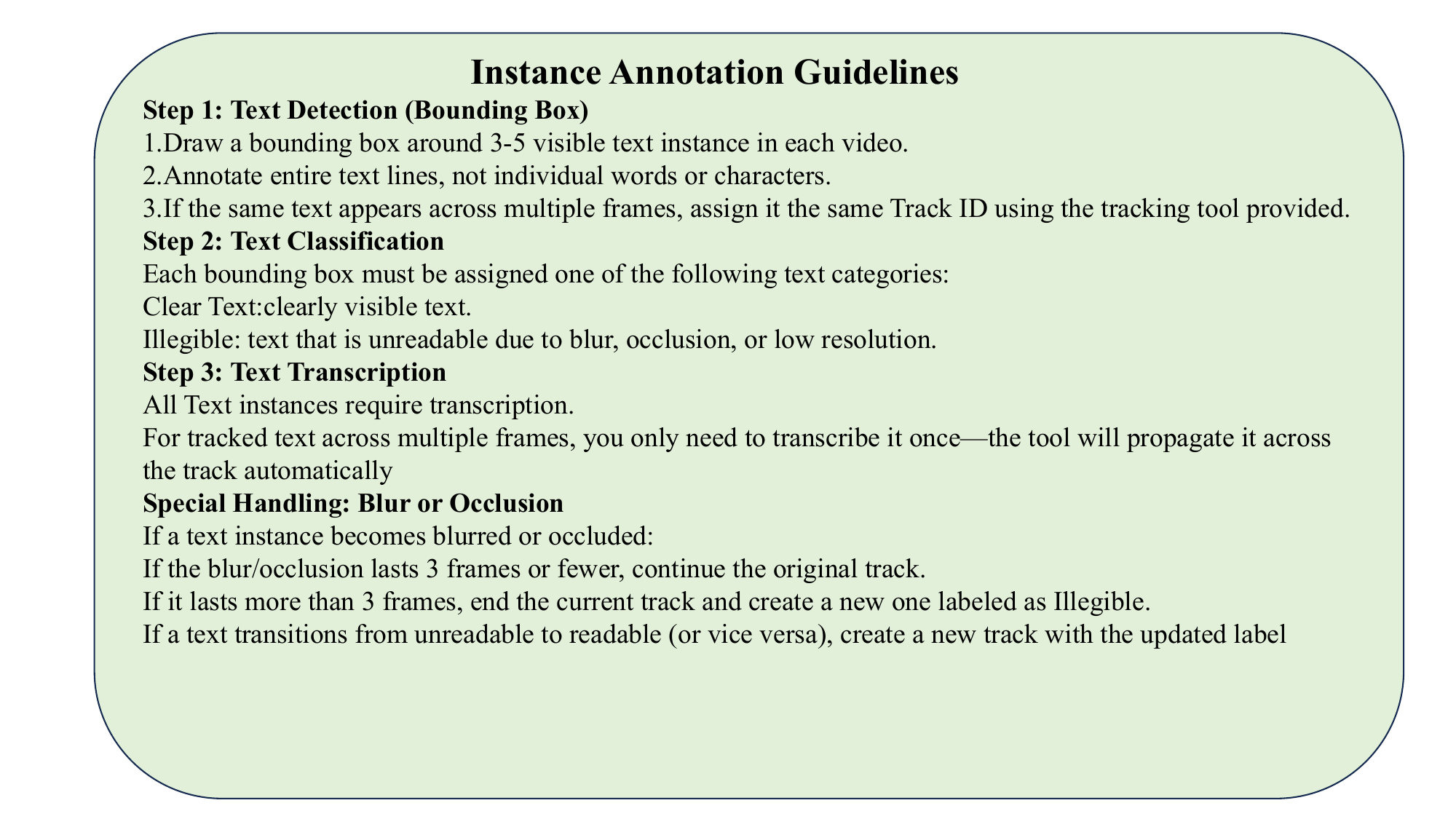}
    \caption{Instance-Level Annotation Guidelines.}
    \label{fig:anno_guide_ins}
\end{figure*}

Each video underwent a two-stage text annotation process. In the first stage, annotators drew tight bounding boxes around visible text lines and assigned each to a category: \textit{ClearText} or \textit{Illegible}. s created. A tracking tool automatically propagated bounding boxes across frames using consistent Track IDs.More Details are shown in Fig.~\ref{fig:anno_guide_ins}.

\subsection*{D.2 Clip-Level Annotation}
Videos shorter than 1 minute were split into 5-second clips; longer ones into 20-second clips. For each clip, annotators recorded all visible, legible text and its temporal span. Repeated instances within a clip were marked only once. Illegible or heavily blurred texts were ignored.
More Details shown in Fig.~\ref{fig:anno_guide_clipV}.

\subsection*{D.3 Video-Level Text Collection}
\begin{figure*}[h]
    \centering
    \includegraphics[width=1\textwidth]{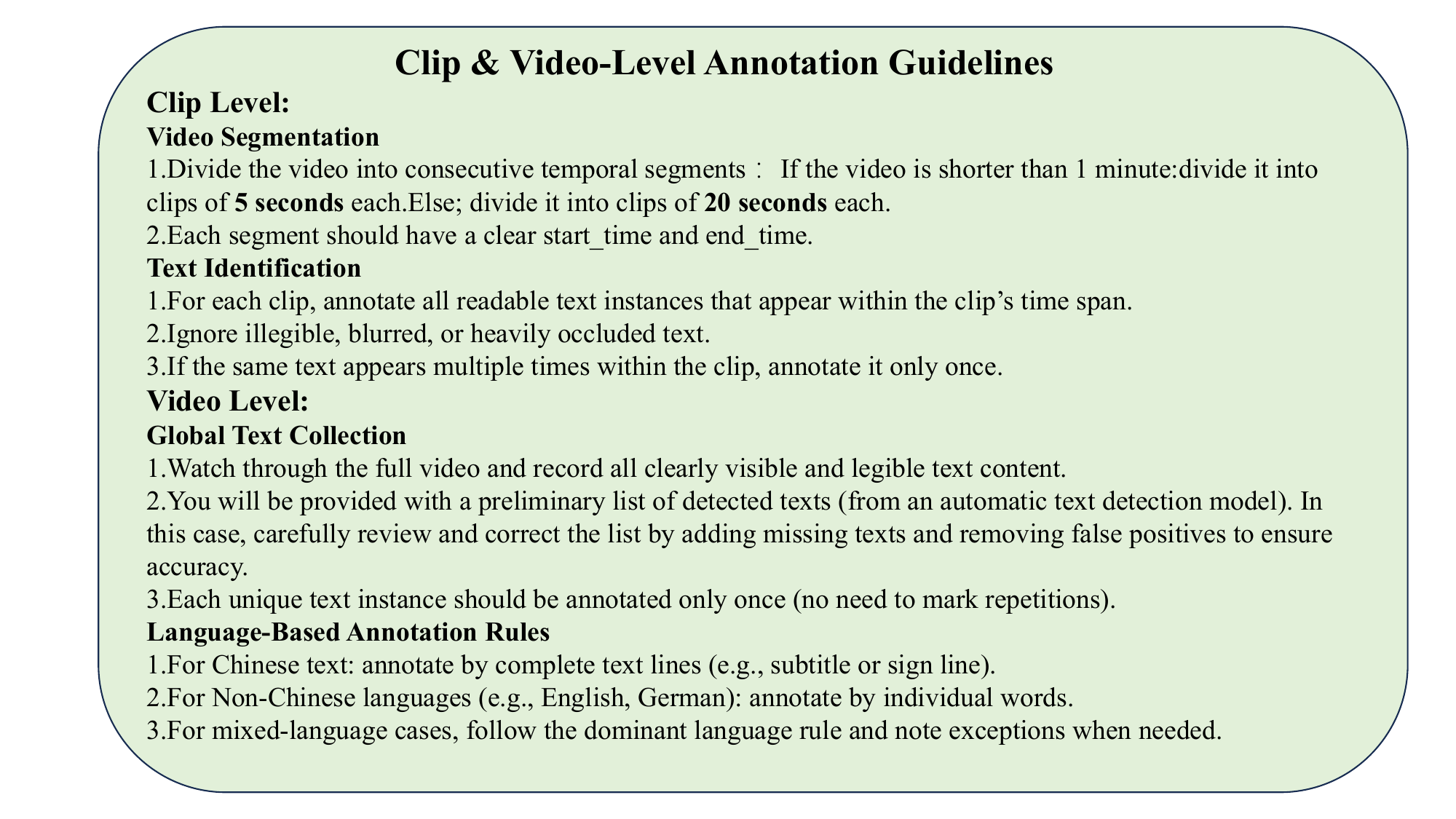}
    \caption{CLip\&Video-Level Annotation Guidelines.}
    \label{fig:anno_guide_clipV}
\end{figure*}

A separate annotation team reviewed the OCR predictions from our model. Annotators removed hallucinated content and added missing instances. Chinese was annotated by full lines; other languages (e.g., English, German) were annotated by words. Each unique string was listed once in the final inventory. More Details shown in Fig.~\ref{fig:anno_guide_clipV}.

\subsection*{D.4 Holistic Reasoning}
\begin{figure*}[h]
    \centering
    \includegraphics[width=1\textwidth]{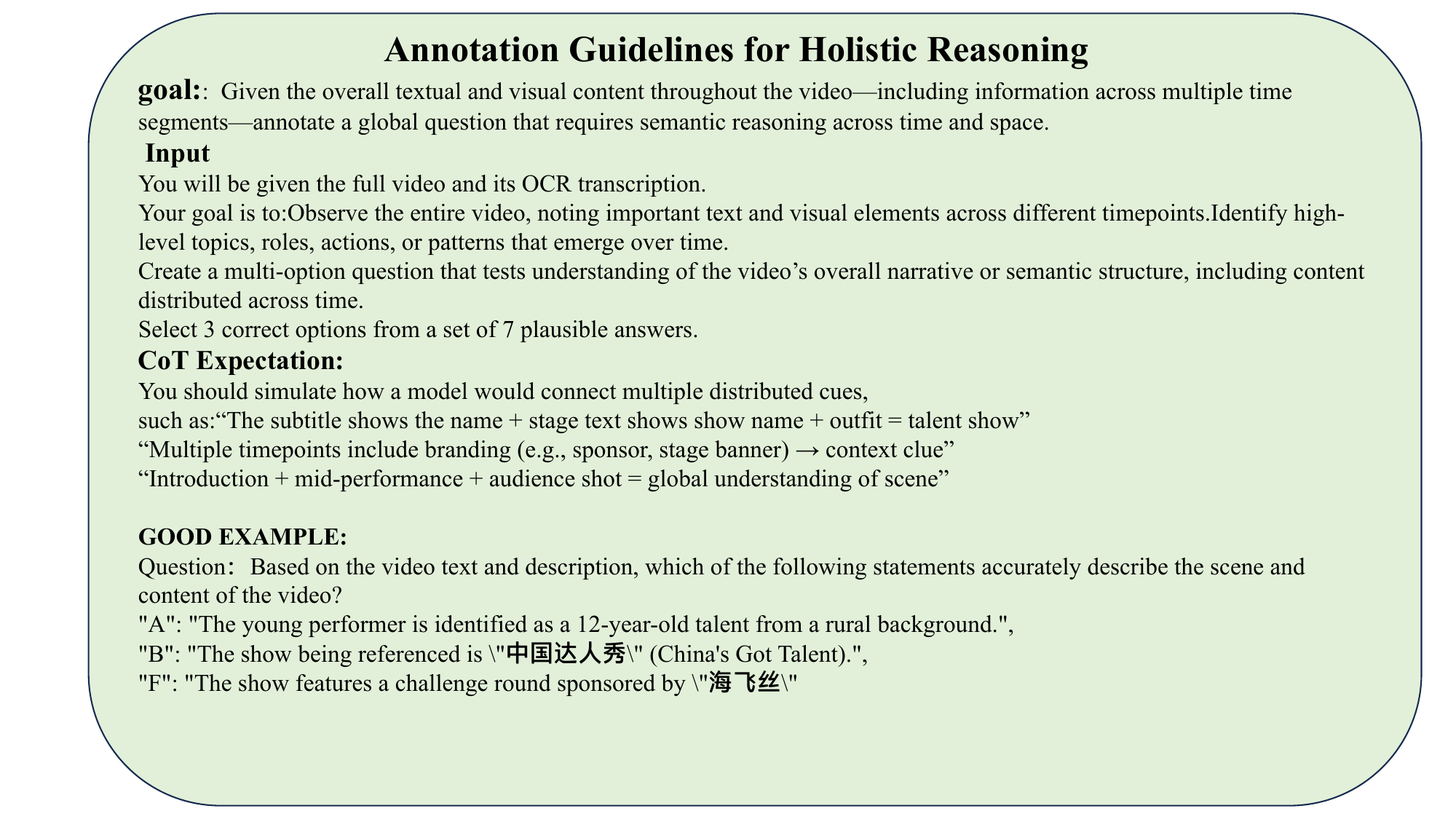}
    \caption{HolisticReasoning Annotation Guidelines.}
    \label{fig:anno_guide_HR}
\end{figure*}

Annotators watched the full video and consulted the video-level text inventory to write one multi-label question per video (see Fig.~\ref{fig:anno_guide_HR}). Each question included seven options describing high-level semantics such as scene, role, topic, or sponsor. 

\subsection*{D.5 Local Reasoning}
\begin{figure*}[h]
    \centering
    \includegraphics[width=1\textwidth]{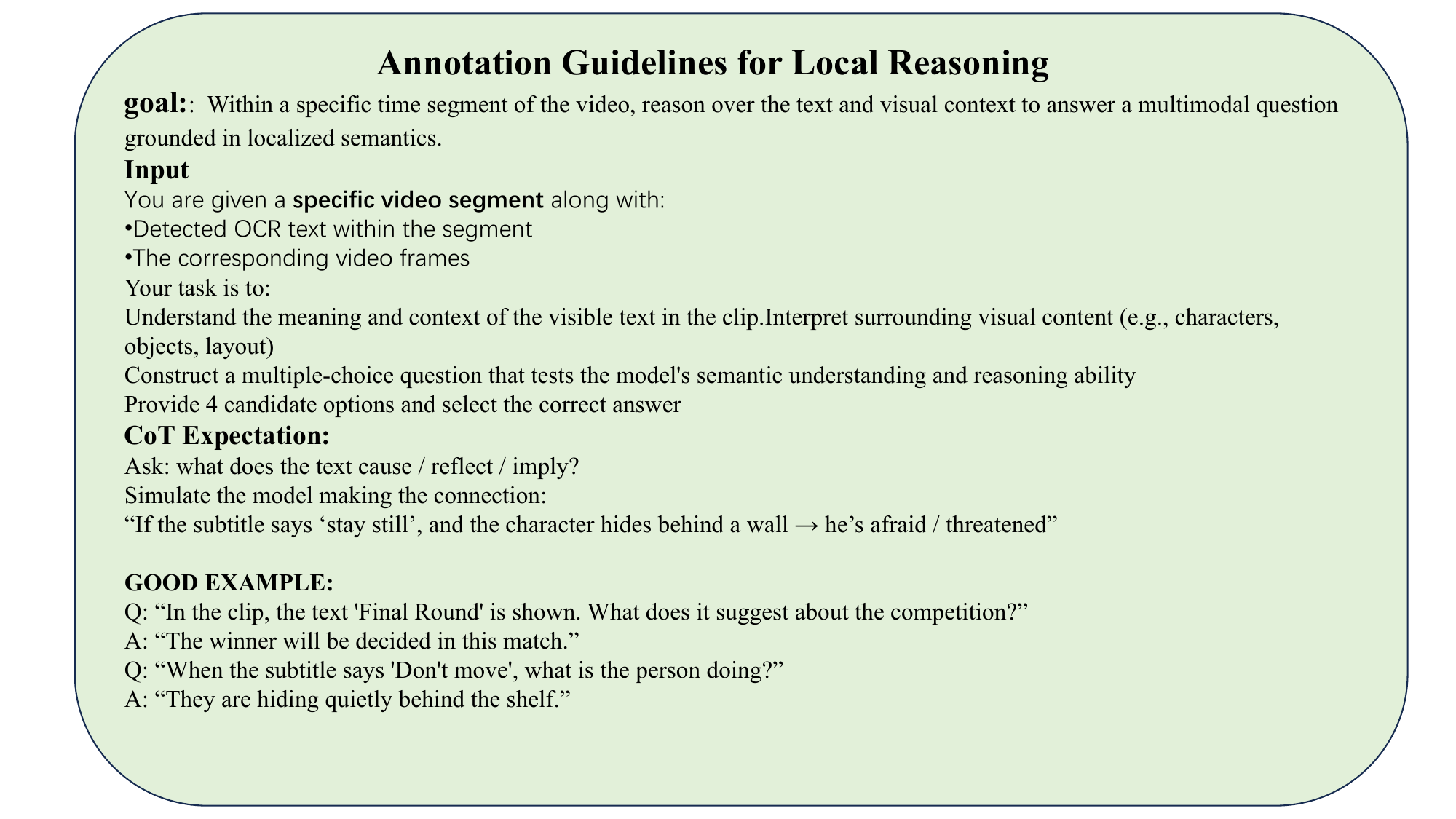}
    \caption{LocalReasoning Annotation Guidelines.}
    \label{fig:anno_guide_LR}
\end{figure*}

For every clip (as defined in D.2), annotators created one four-option multiple-choice question requiring reasoning between localized text and visual context (e.g., subtitle  character behavior). The question must require multimodal reasoning and not be solvable using text or image alone. More Details shown in Fig.~\ref{fig:anno_guide_LR}.

\subsection*{D.6 Temporal Causal Reasoning (TCR)}
\begin{figure*}[h]
    \centering
    \includegraphics[width=1\textwidth]{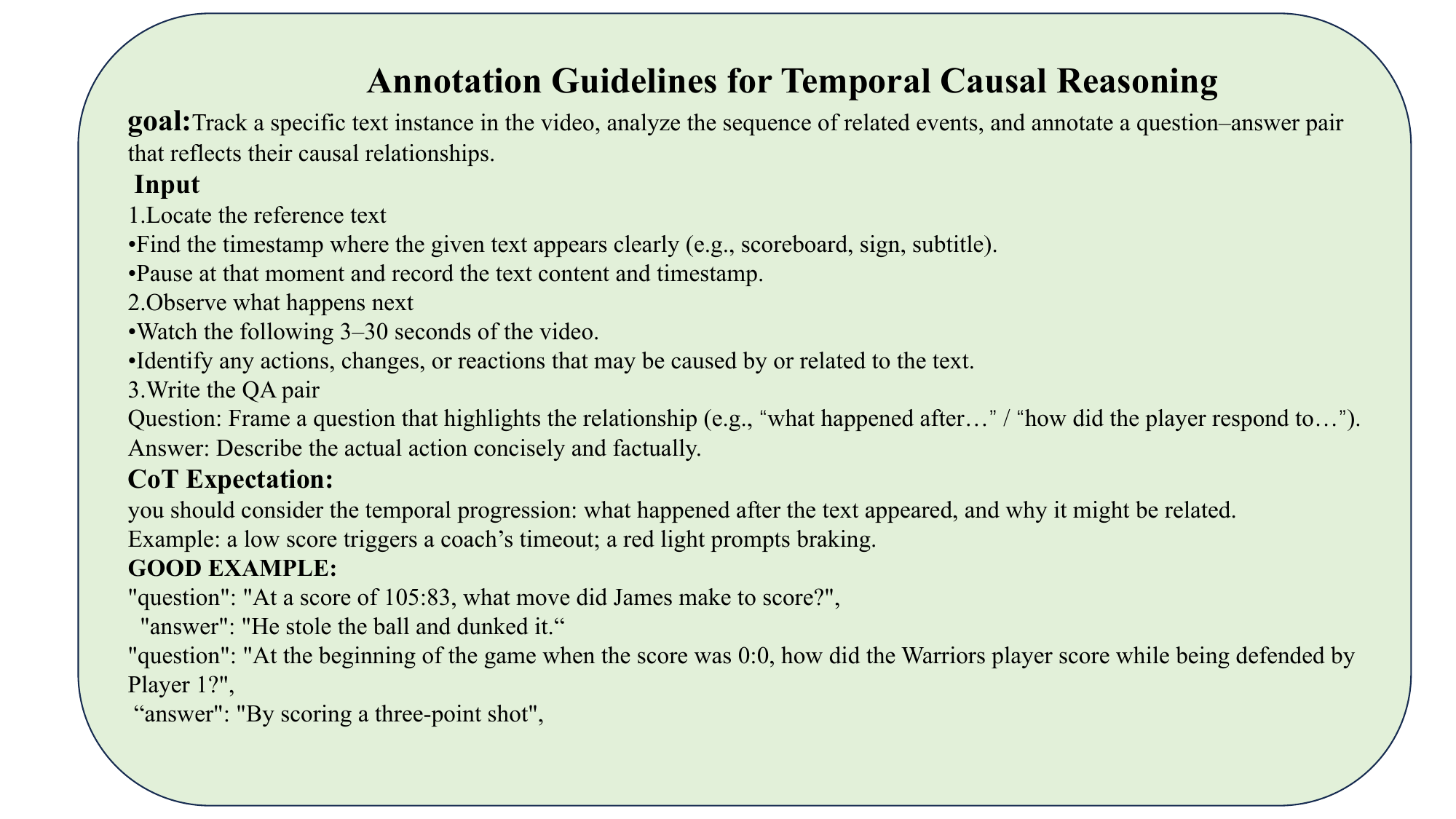}
    \caption{TemporalCausalReasoning Annotation Guidelines.}
    \label{fig:anno_guide_TCR}
\end{figure*}

Given a reference text (e.g., scoreboard or subtitle), annotators identified the timestamp of its appearance, observed the following 3--30 seconds, and formulated a causal reasoning question. The answer was a single factual sentence describing the resulting action. Each QA pair was anchored to the cue’s timestamp. More Details shown in Fig.~\ref{fig:anno_guide_TCR}.

\subsection*{D.7 Spatial Reasoning (SR)}
\begin{figure*}[h]
    \centering
    \includegraphics[width=1\textwidth]{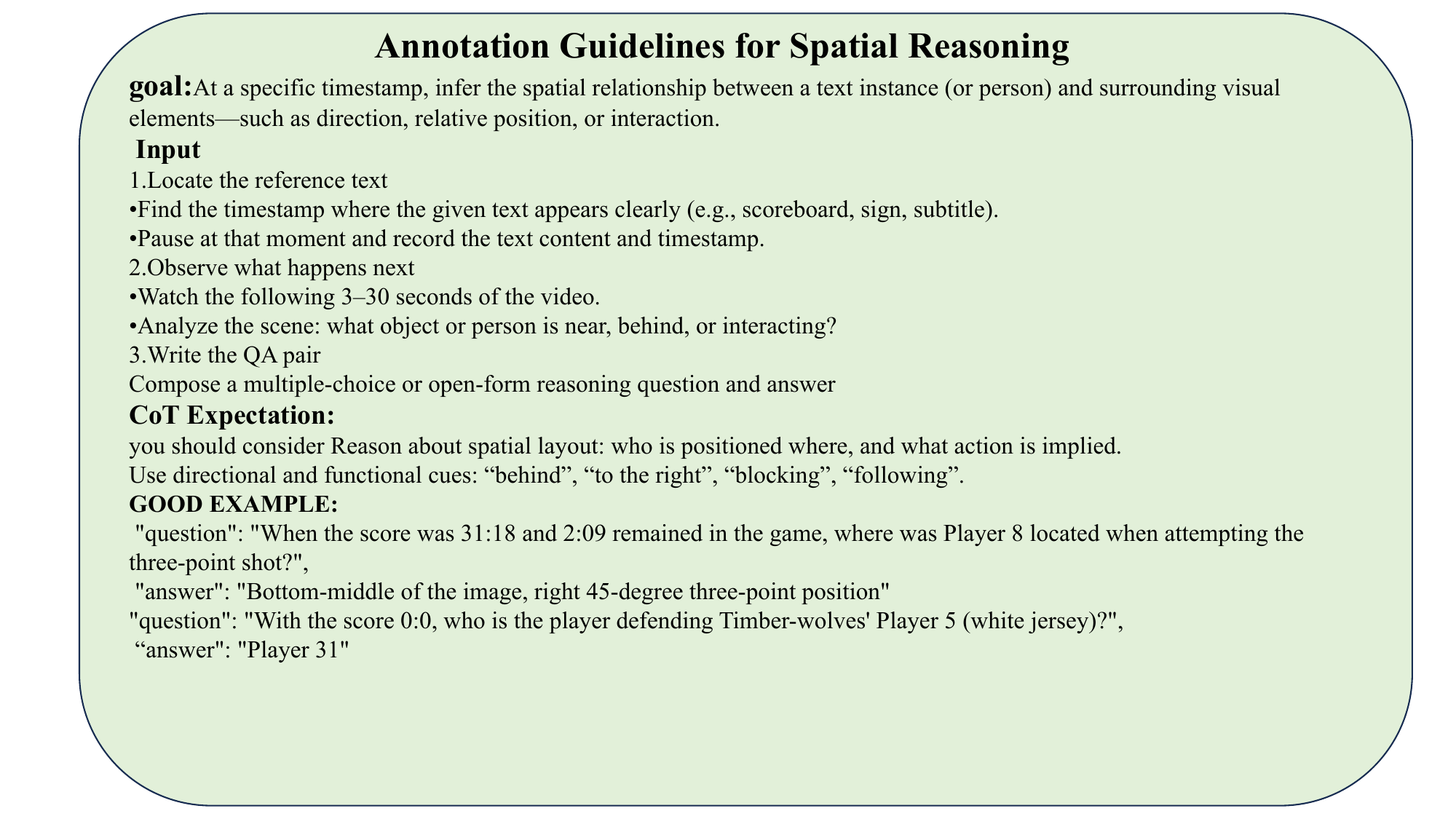}
    \caption{SpatialReasoning Annotation Guidelines.}
    \label{fig:anno_guide_SR}
\end{figure*}

As shown in Fig.~\ref{fig:anno_guide_SR}, at a given timestamp, annotators located a reference text or entity and constructed a question requiring reasoning over its spatial relation to nearby visual elements (e.g., direction, proximity, interaction).

\paragraph{Quality Control} All annotations underwent double review. Each item was cross-validated by a second annotator, and disagreements were resolved by expert adjudication. On a random sample of 200 items, we achieved an average inter-annotator agreement of \textbf{0.81} (Cohen’s~), indicating high reliability.

\section{Details of Experimental Settings}
\label{appendix:settings}
\label{sub-appendix:evaluation-baselines}

\subsection*{E.1 Model Configuration}

In this section, we outline the primary baselines evaluated on our VidText.To ensure fair comparison across both open- and closed-source models, we explicitly standardize frame sampling and spatial resolution for each baseline as summarized in Tab.~\ref{tab:baseline_cfg}.

For proprietary models such as GPT-4o, Gemini 1.5 (Pro and Flash), and GPT-4-Turbo, we follow their official or API-supported settings. \texttt{GPT-4o} models support up to approximately 500 images inputs, for which we adopt a uniform sampling rate of 0.5 fps with an input resolution of $512 \times 512$  to accommodate most of our videos. \texttt{GPT-4-Turbo} is restricted to 16 frames, uniformly sampled across the video, and resized to the same resolution.

For open-source models, we align each configuration with their original public implementations. \texttt{VideoChat-Flash}, \texttt{Qwen2-VL (7B)}, and All \texttt{Qwen2.5-VL} variants (3B/7B/72B) operate under a 1~fps sampling strategy, with a maximum of 768 frames extracted per video. Models that support extended temporal contexts—such as \texttt{VideoLLaMA 3}, \texttt{InternVL 2.5}, and \texttt{LLaVA-OV}—are provided with 64 uniformly sampled frames, resized to $336 \times 336$. \texttt{ShareGPT4Video} also uses 64 frames, but with a reduced spatial resolution of $224 \times 224$. \texttt{LongVU} and \texttt{LongVA}, are evaluated with sparse and extended frame settings. \texttt{LongVU} uses 1~fps sampling, while \texttt{LongVA} accepts up to 128 uniformly distributed frames. \texttt{MiniCPM-V2.6} applies a fixed 64-frame sliding window, following its official implementation.

%------------------------------------------------------------
%  Baseline Configuration: sampling & spatial resolution only
%------------------------------------------------------------
\begin{table*}[t]
  \centering
  \small
  \setlength\tabcolsep{6pt}
  \caption{Frame–sampling and input‐resolution settings for baselines.}
  \label{tab:baseline_cfg}
  \begin{tabular}{lccc}
    \toprule
    \textbf{Model} & \textbf{Size} & \textbf{Sampling} & \textbf{Resolution} \\
    \midrule
    \rowcolor[HTML]{F1F6EC}\multicolumn{4}{l}{\textit{\textbf{Proprietary MLLMs}}} \\
    \rowcolor[HTML]{F1F6EC} GPT-4-Turbo               & –    & 16 frames   & $512^{2}$ \\
    \rowcolor[HTML]{F1F6EC} Gemini 1.5 Flash          & –    & 1 fps       & $512^{2}$ \\
    \rowcolor[HTML]{F1F6EC} GPT-4o                    & –    & 0.5 fps       & $512^{2}$ \\
    \rowcolor[HTML]{F1F6EC} Gemini 1.5 Pro            & –    & 1 fps       & $512^{2}$ \\
    \midrule
    \rowcolor[HTML]{E3F8F8}\multicolumn{4}{l}{\textit{\textbf{Open-source MLLMs}}} \\
    \rowcolor[HTML]{E3F8F8} LongVU                    & 3 B  & 1 fps       & $448^{2}$ \\
    \rowcolor[HTML]{E3F8F8} Qwen2.5-VL                & 3 B  & 1 fps       & $448^{2}$ \\
    \rowcolor[HTML]{E3F8F8} Video-XL-Pro              & 7 B  & 1 fps       & $448^{2}$ \\
    \rowcolor[HTML]{E3F8F8} LongVA                    & 7 B  & 128 frames  & –         \\
    \rowcolor[HTML]{E3F8F8} MiniCPM-V2.6              & 7 B  & 64 frames   & $448^{2}$ \\
    \rowcolor[HTML]{E3F8F8} VideoChat-Flash           & 7 B  & 1 fps       & $448^{2}$ \\
    \rowcolor[HTML]{E3F8F8} Qwen2-VL                  & 7 B  & 1 fps       & $448^{2}$ \\
    \rowcolor[HTML]{E3F8F8} Qwen2.5-VL                & 7 B  & 1 fps       & $448^{2}$ \\
    \rowcolor[HTML]{E3F8F8} VideoLLaMA 3              & 7 B  & 64 frames   & $336^{2}$ \\
    \rowcolor[HTML]{E3F8F8} ShareGPT4Video            & 8 B  & 64 frames   & $224^{2}$ \\
    \rowcolor[HTML]{E3F8F8} Oryx-1.5                  & 32 B & 64 frames   & $336^{2}$ \\
    \rowcolor[HTML]{E3F8F8} LLaVA-OV                  & 72 B & 64 frames   & $336^{2}$ \\
    \rowcolor[HTML]{E3F8F8} Qwen2.5-VL                & 72 B & 1 fps       & $448^{2}$ \\
    \rowcolor[HTML]{E3F8F8} InternVL 2.5              & 78 B & 64 frames   & $336^{2}$ \\
    \bottomrule
  \end{tabular}
\end{table*}

\subsection*{E.2 Human Performance Study}

To assess the upper-bound of performance on \textsc{VidText}, we conducted a controlled human evaluation across all tasks in our benchmark. Three annotators with experience in video analysis and text recognition were recruited to answer a representative subset of questions spanning all eight task types. Each participant was given access to the full video content and instructed to answer using their best judgment, without time constraints. The average human accuracy across all tasks reaches \textbf{89.5\%}, substantially outperforming all evaluated models. In particular, humans demonstrated near-perfect scores in holistic and local OCR, reasoning, and spatial understanding tasks, highlighting the gap between human-level comprehension and the capabilities of current multimodal large models. These results serve as a reference ceiling for future model development and underline the complexity and nuance of the video-text understanding challenges posed by \textsc{VidText}.More details are shown in Tab.~\ref{tab:main_result}.

\subsection*{E.3 Experiment Environment.}
All experiments are conducted on a server equipped with 4$\times$NVIDIA A100 GPUs (80GB each). Model inference and evaluation are implemented in PyTorch with mixed-precision support. 

\section{Model Prompts}
\label{appendix:prompts}

\begin{figure*}[h]
    \centering
    \includegraphics[width=1\textwidth]{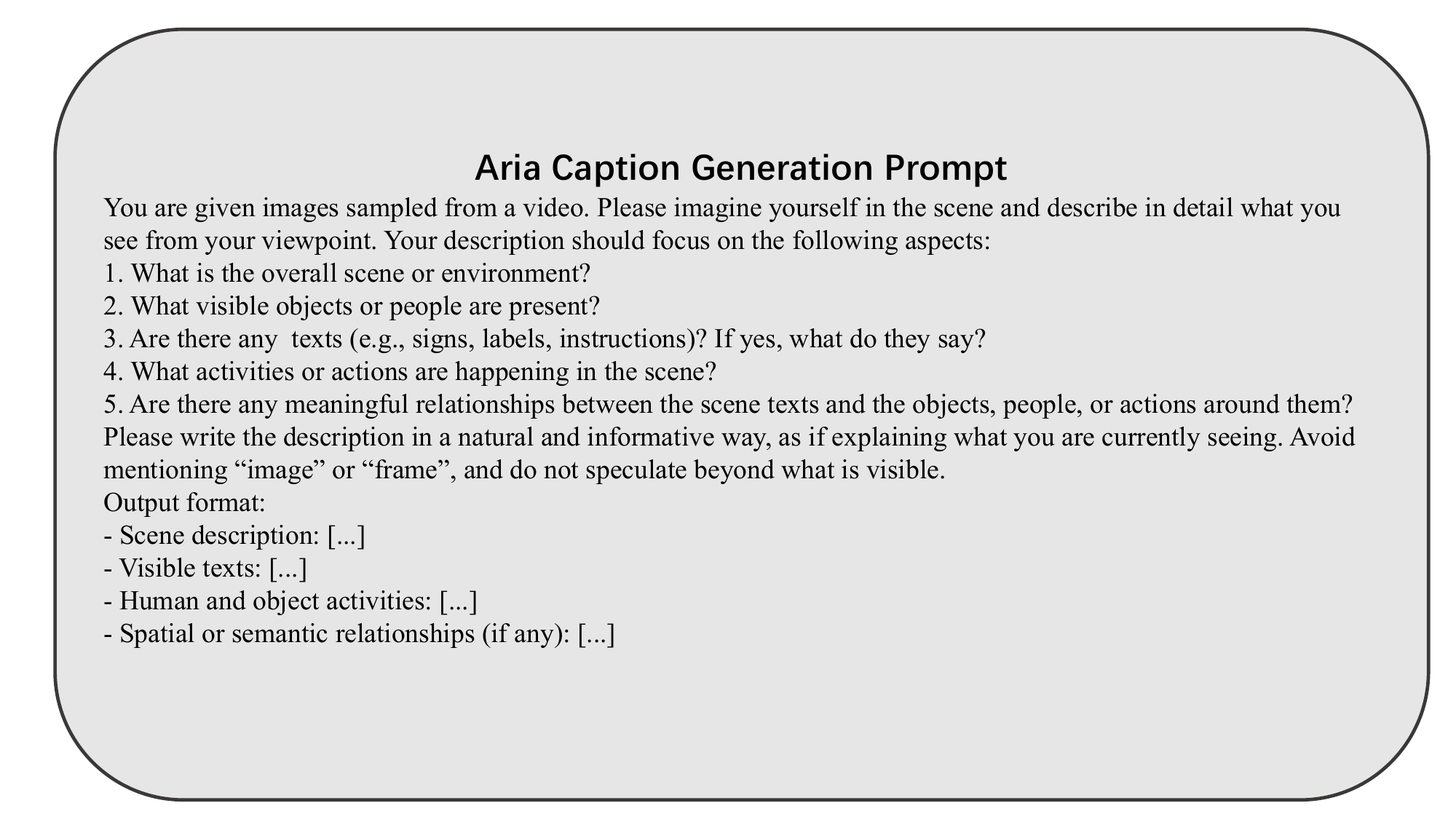}
    \caption{Prompt template used for Aria to generate frame-level captions.}
    \label{fig:Aria_promp}
\end{figure*}

Fig.~\ref{fig:Aria_promp} shows the prompt template used to obtain detailed frame-level captions from the Aria model. The prompt includes instructions to describe the scene, detect visible text, summarize actions, and relate them spatially and semantically. Tab.~\ref{tab:prompt-templates} lists the standardized prompt templates used for each task in VidText.

\begin{table*}[t]
\centering
\caption{Prompt templates used for VidText tasks.}
\label{tab:prompt-templates}
\renewcommand{\arraystretch}{1.3}
\resizebox{\textwidth}{!}{%
\begin{tabular}{ll}
\toprule
\textbf{Task} & \textbf{Prompt Template} \\
\midrule
\textbf{Holistic OCR} & \begin{tabular}[c]{@{}l@{}}"Recognize all visual texts in the video.\\
If the text is not in English, do not provide an English translation.\\
Do not include any descriptions, narrative, or context.\\
Output only the extracted text lines, each on a new line."\end{tabular} \\
\midrule
\textbf{Holistic Reasoning} & \begin{tabular}[c]{@{}l@{}}"Watch the video carefully and select the correct three answers.\\
Question: \{question\}\\
Options: \{options\}\\
Please output your answer in the format: \texttt{Correct Answers: A, B, C}"\end{tabular} \\
\midrule
\textbf{Local OCR} & \begin{tabular}[c]{@{}l@{}}"Watch the video and answer the following question based on its content.\\
Question: \{question\}\\
Please output only the texts that appear in the specified time interval as a JSON array of strings,\\
with each element representing one piece of text. Do not include any additional description or translation."\end{tabular} \\
\midrule
\textbf{Local Reasoning} & \begin{tabular}[c]{@{}l@{}}"Watch the video and answer the following multiple-choice question based on its content.\\
Question: \{question\}\\
Options:\\
Option A: \{text\}\\
Option B: \{text\}\\
...\\
Please select the correct option."\end{tabular} \\
\midrule
\textbf{Text Localization} & \begin{tabular}[c]{@{}l@{}}"Watch the video and answer the following question based on its content.\\
Please provide the time interval (in seconds, precise to 0.1s) during which the text appears in the video.\\
Output your answer in JSON format with keys 'start' and 'end'. For example: \{\texttt{"start": 0.0, "end": 30.0}\}.\\
Do not include any extra commentary."\end{tabular} \\
\midrule
\textbf{Temporal Causal Reasoning} & \begin{tabular}[c]{@{}l@{}}"Watch the video and answer the following multiple-choice question based on its content.\\
Question: \{question\}\\
Options:\\
Option A: \{text\}\\
Option B: \{text\}\\
...\\
Please select the correct option."\end{tabular} \\
\midrule
\textbf{Text Tracking} & \textit{(Same prompt as Spatial Reasoning)} \\
\midrule
\textbf{Spatial Reasoning} & \begin{tabular}[c]{@{}l@{}}"Watch the video and answer the following multiple-choice question based on its content.\\
Question: \{question\}\\
Options:\\
Option A: \{text\}\\
Option B: \{text\}\\
...\\
Please select the correct option."\end{tabular} \\
\bottomrule
\end{tabular}
}
\end{table*}

\begin{figure}[t]
    \centering
    \includegraphics[width=0.9\linewidth]{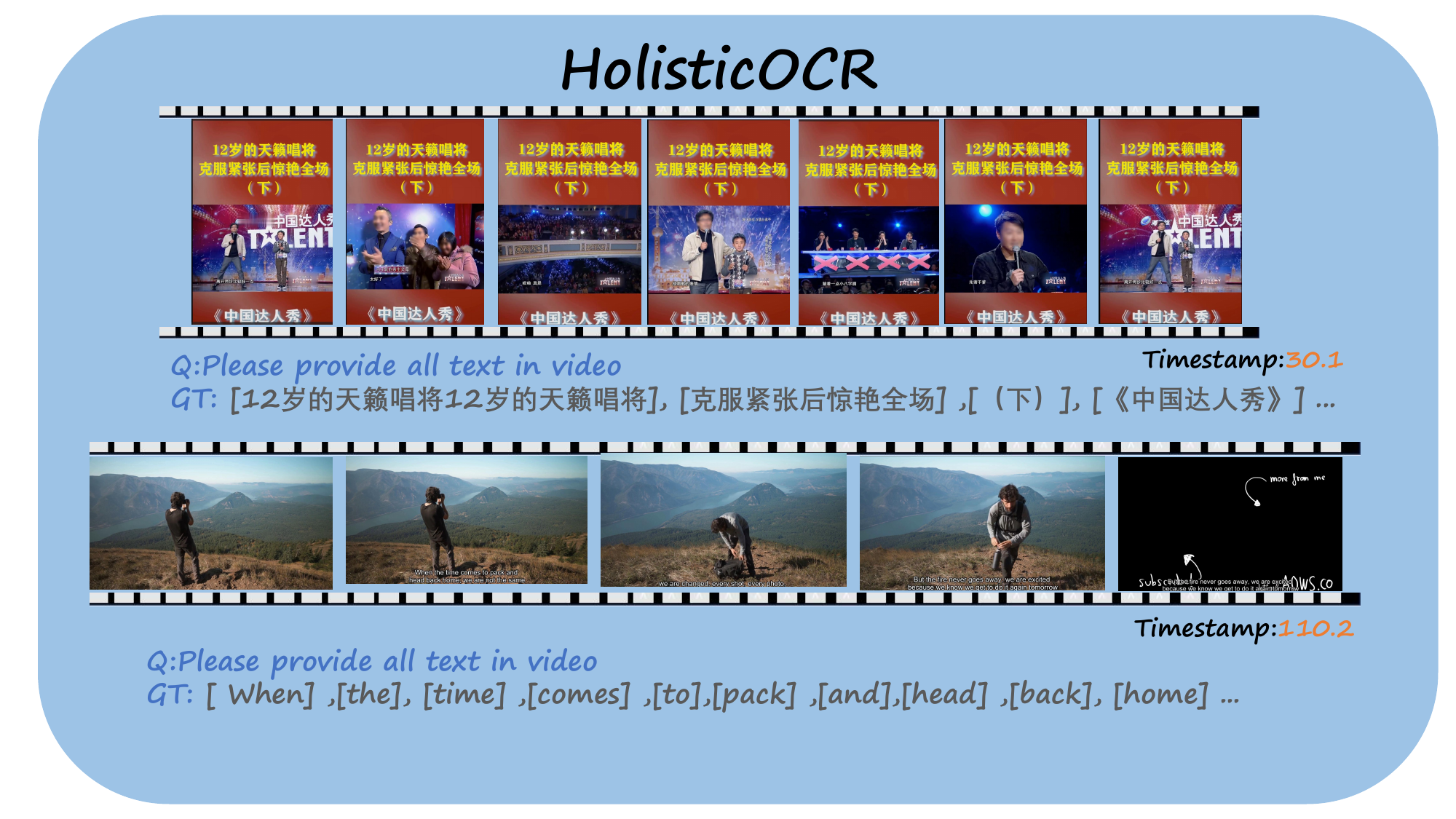}
   % \vspace{1pt}
    \includegraphics[width=0.9\linewidth]{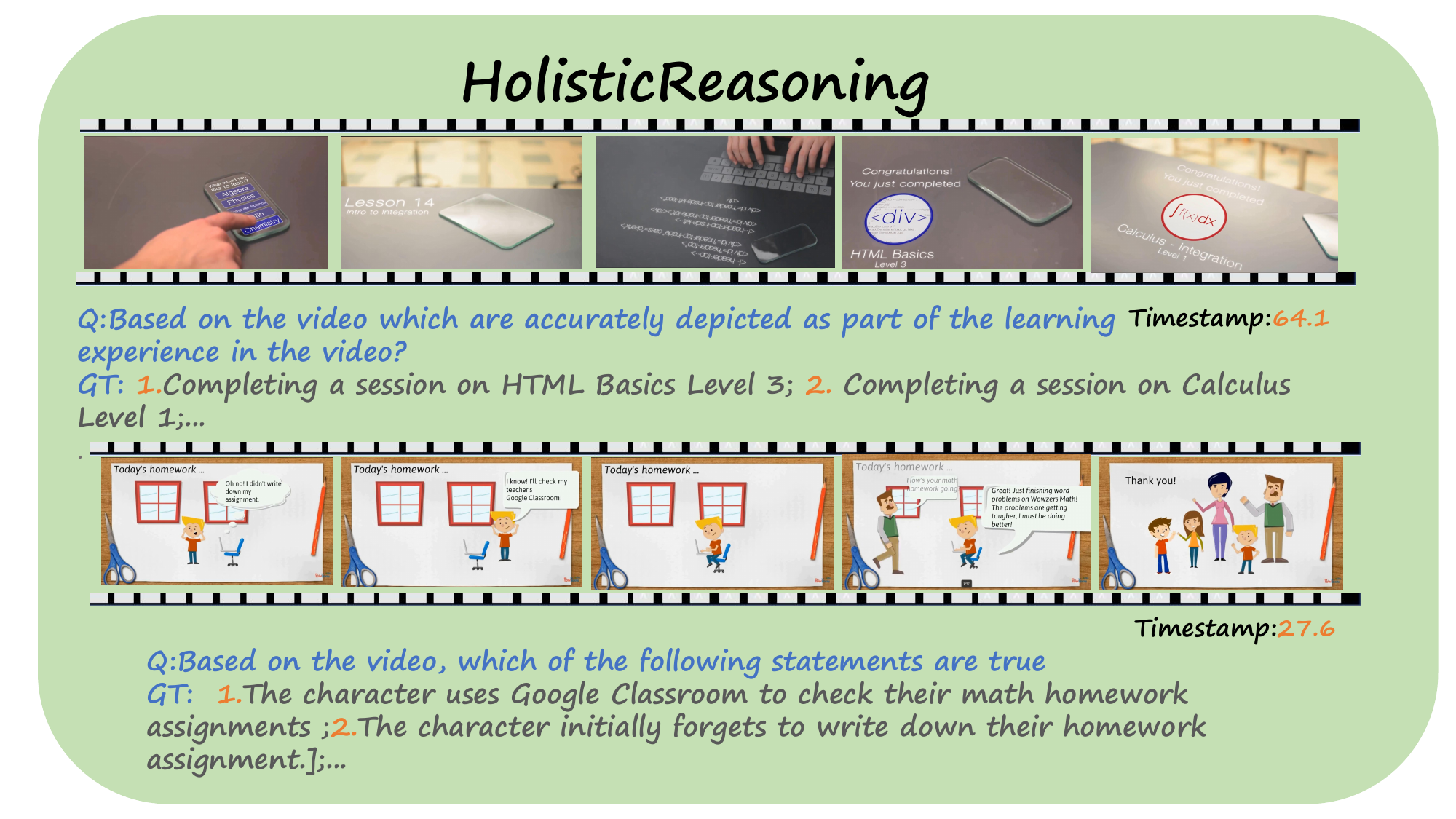}
    %\vspace{1pt}
    \includegraphics[width=0.9\linewidth]{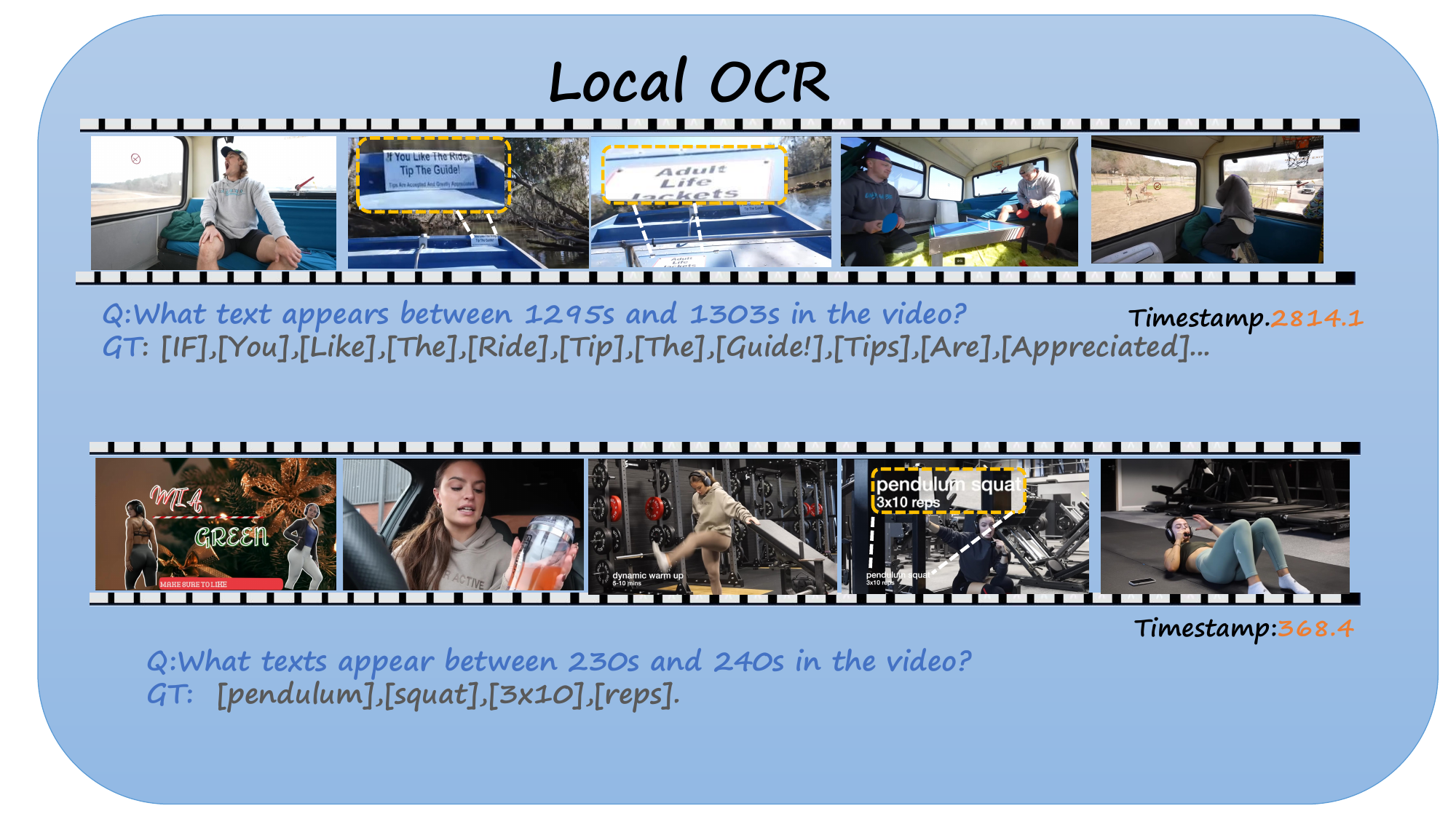}
    \caption{(Top) More examples of HolisticOCR. (Middle) More examples of HolisticReasoning. (Bottom) More examples of LocalOCR.}
    \label{fig:more_HR_HOCR_lOCR}
\end{figure}

\begin{figure}[t]
    \centering
    \includegraphics[width=0.9\linewidth]{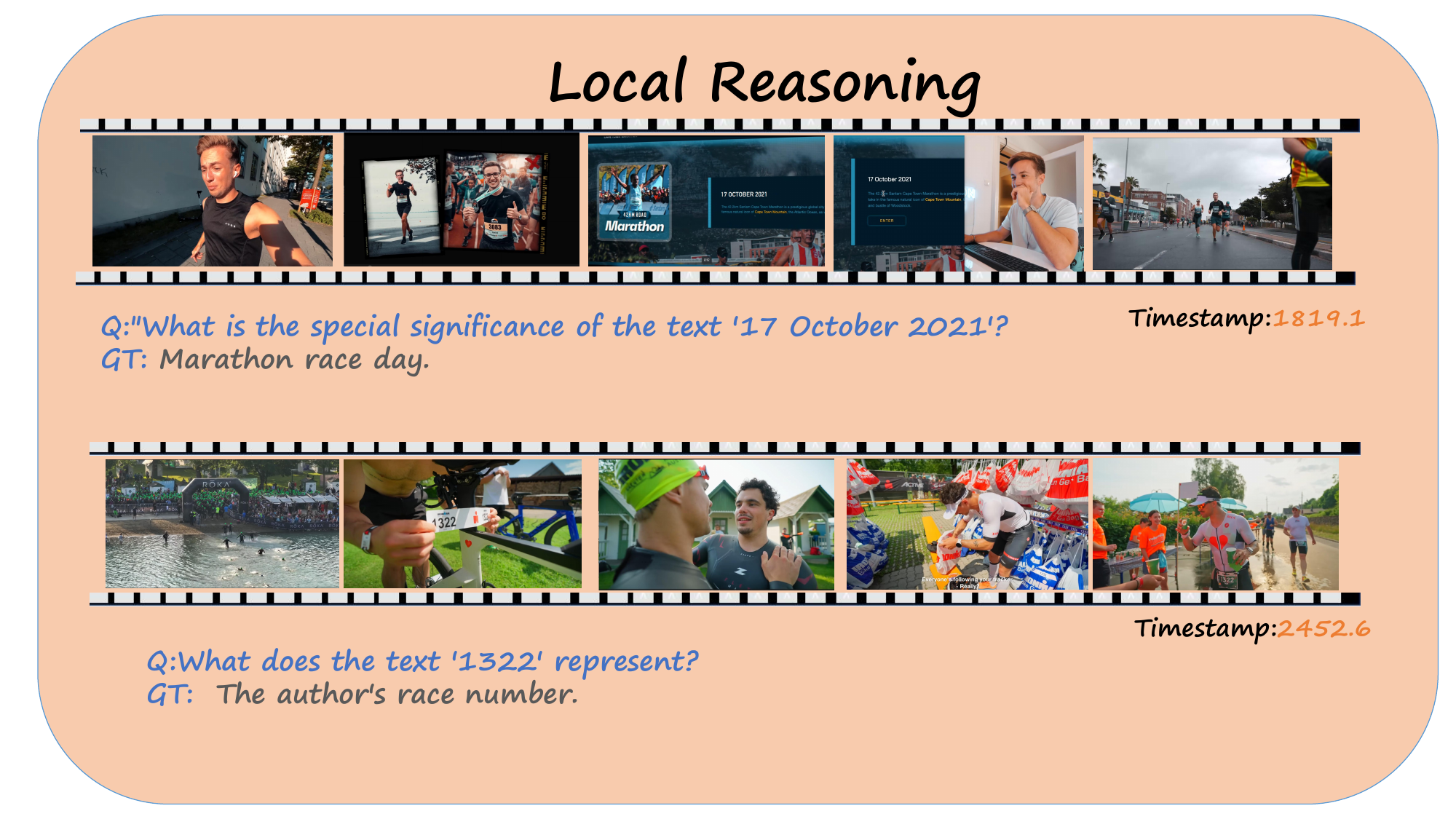}
    %\vspace{4pt}
    \includegraphics[width=0.9\linewidth]{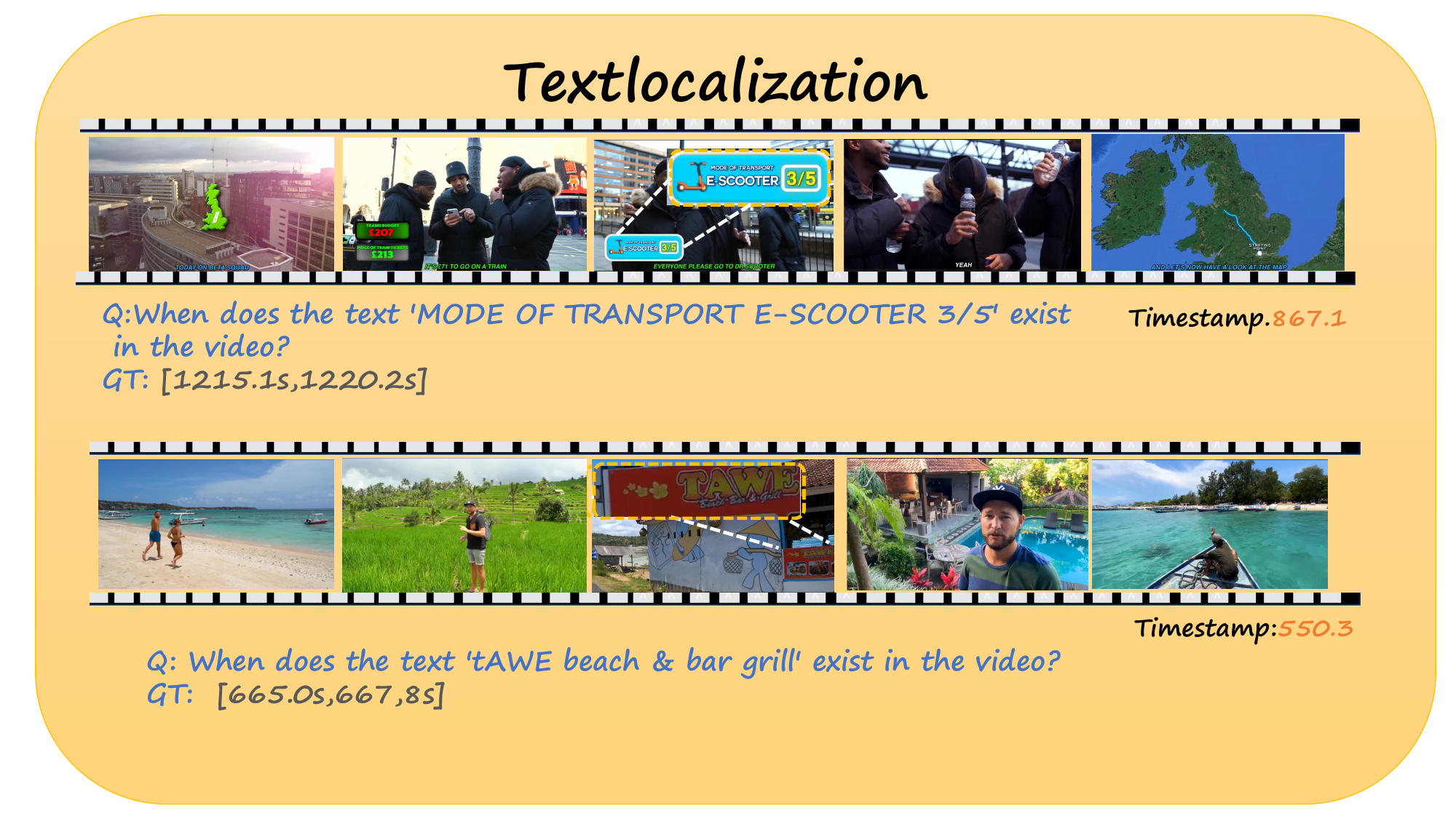}
   % \vspace{4pt}
    \includegraphics[width=0.9\linewidth]{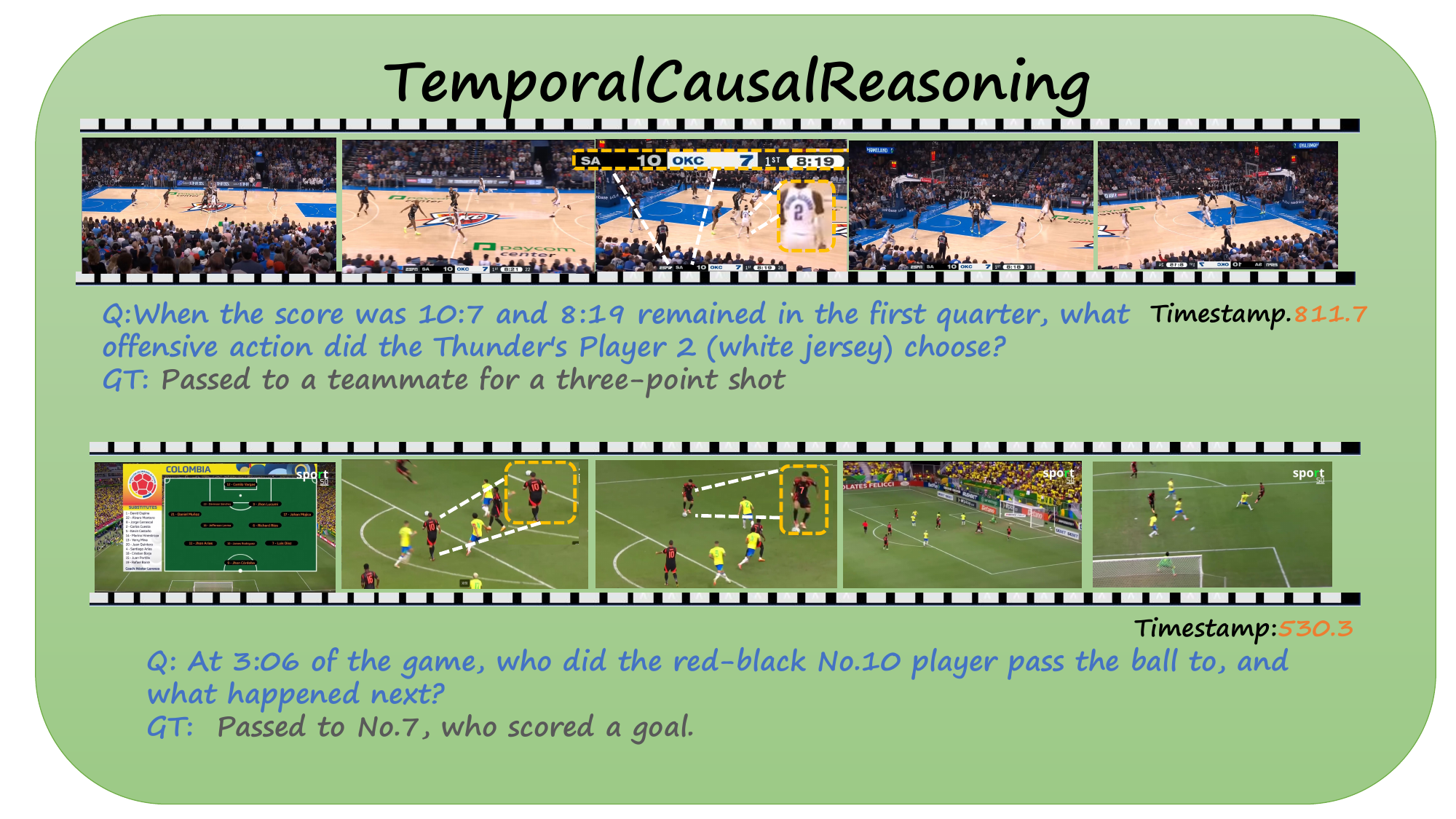}
    \caption{(Top) More examples of LocalReasoning. (Middle) More examples of TextLocalization. (Bottom) More examples of TemporalCausalReasoning.}
    \label{fig:more_LR_TL_TCR}
\end{figure}

\clearpage

\section{More Visualization Results}
\label{appendix:vis}

\begin{figure}[t]
    \centering
    \includegraphics[width=0.9\linewidth]{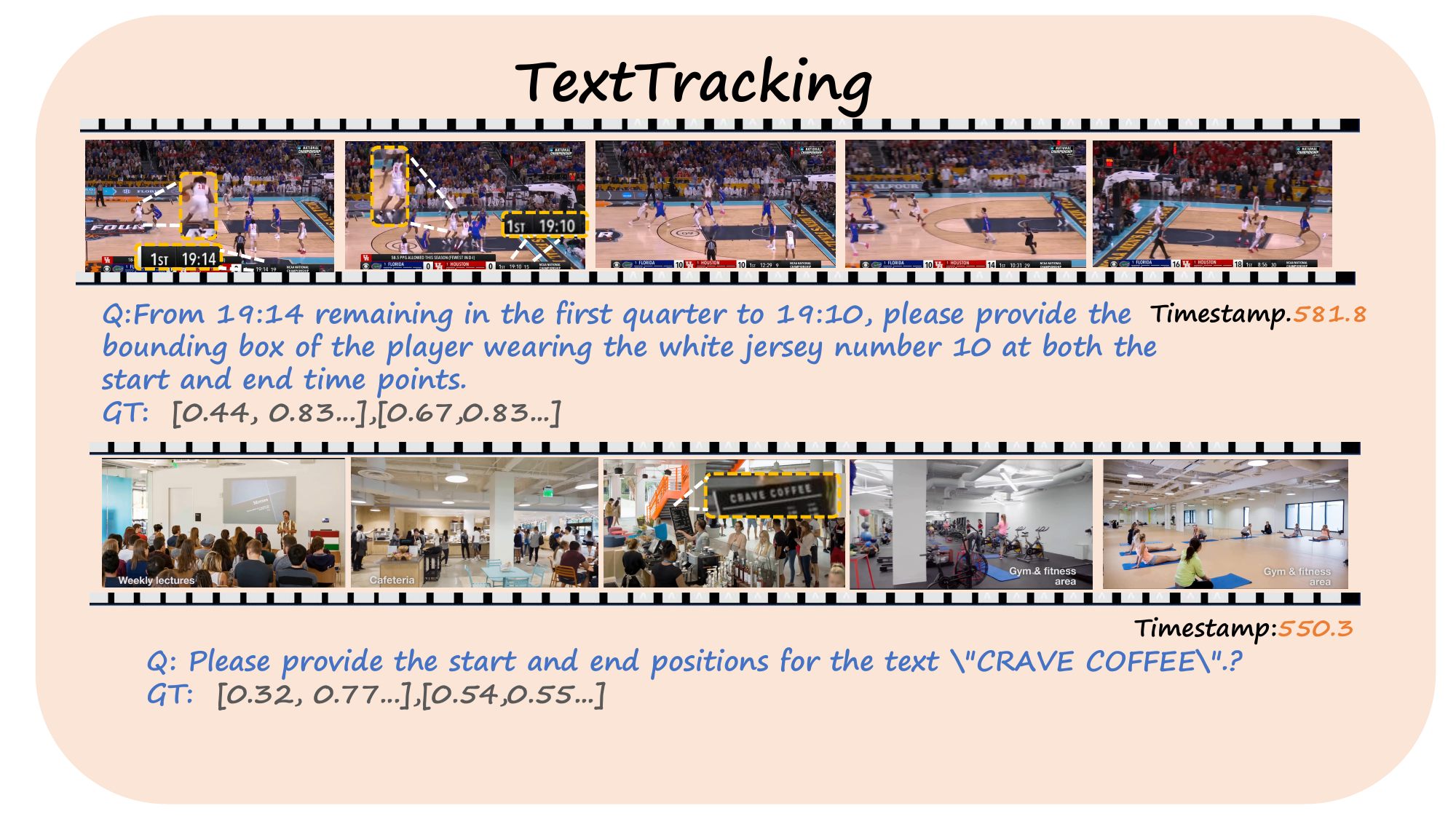}

    \includegraphics[width=0.9\linewidth]{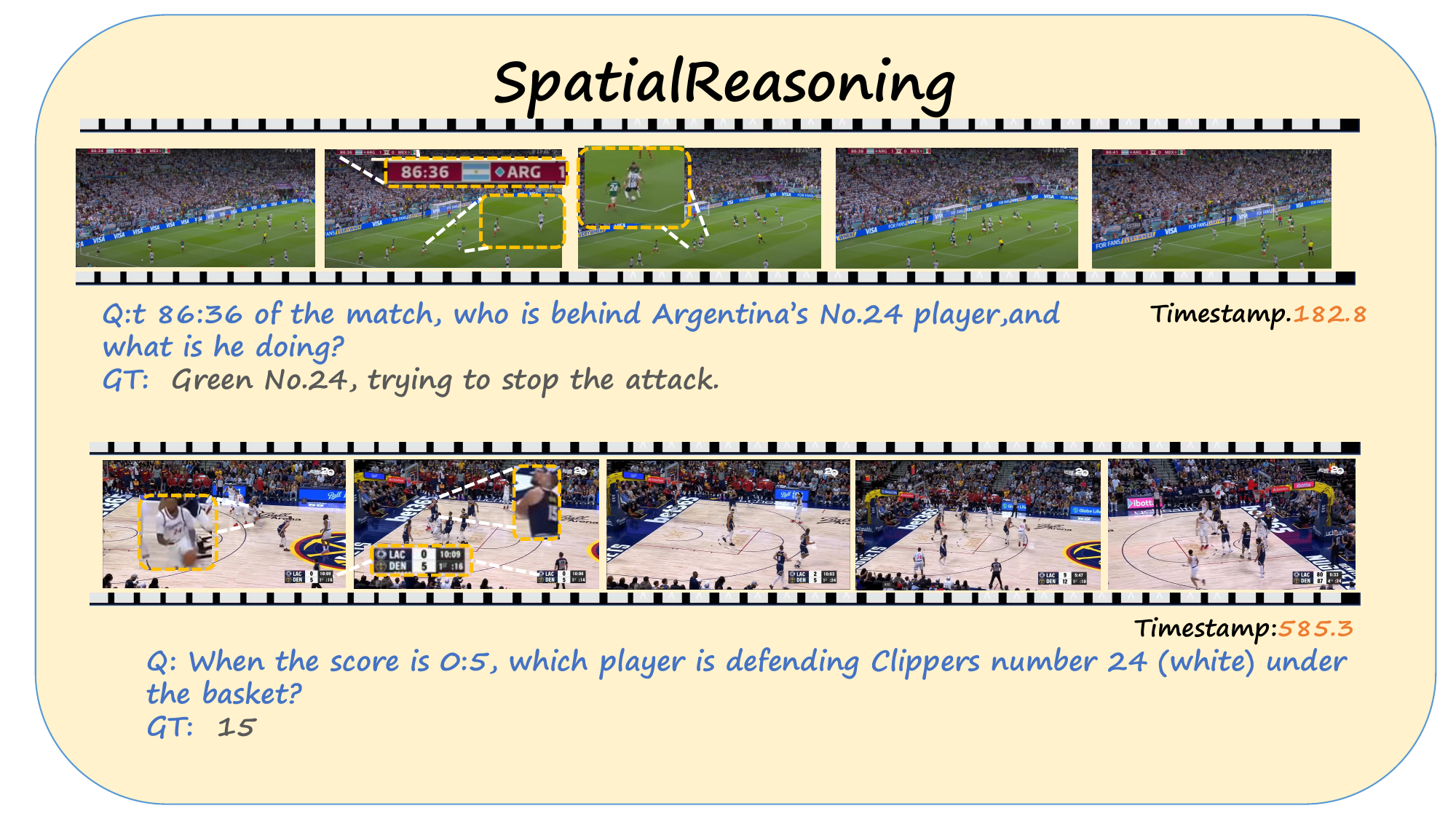}
    \caption{(Top) More examples of TextTracking. (Bottom) More examples of SpatialReasoning.}
    \label{fig:more_TT_SR}
\end{figure}

%%%%%%%%%%%%%%%%%%%%%%%%%%%%%%%%%%%%%%%%%%%%%%%%%%%%%%%%%%%%
We present additional visualizations of our VidText annotation examples in Fig~\ref{fig:more_HR_HOCR_lOCR}, 
\ref{fig:more_LR_TL_TCR}, and \ref{fig:more_TT_SR}.

\end{document}